\documentclass[preprint,12pt]{elsarticle}
\newdefinition{remark}{Remark}
\newdefinition{definition}{Definition}%
\newdefinition{condition}{Condition}
\newdefinition{assumption}{Assumption}
\newdefinition{lemma}{Lemma}
\newdefinition{subsublemma}{Lemma C.}
\newdefinition{appendixdefinition}{Definition}[lemma]%
\newdefinition{sublemma}[appendixdefinition]{Lemma}%
\newproof{proof}{Proof}
\usepackage[T1]{fontenc}
\usepackage[utf8]{inputenc}
\usepackage{amsmath}
\usepackage{physics}
\usepackage{xcolor}
\usepackage{hyperref}
\usepackage{algorithm}
\usepackage{url}
\usepackage{etoolbox}
\usepackage{multirow}
\usepackage{algpseudocode}
\usepackage{mathtools}
\usepackage{graphicx}
\usepackage{amsmath, amssymb}
\usepackage{subcaption}
\usepackage{dsfont}

\usepackage{amsthm}
\usepackage{slashbox}
\DeclareMathOperator*{\argmax}{argmax}

\DeclareMathOperator*{\supremum}{sup}

\journal{Neurocomputing}
\begin{document}
\renewcommand{\arraystretch}{1.5}
\begin{frontmatter}



\title{NeuralBO: A Black-box Optimization Algorithm using Deep Neural Networks}


\author[1]{Dat Phan-Trong\corref{cor1}}
\ead{trongp@deakin.edu.au}

\author[1]{Hung Tran-The}
\ead{hung.tranthe@deakin.edu.au}

\author[1]{Sunil Gupta}
\ead{sunil.gupta@deakin.edu.au}

\cortext[cor1]{Corresponding author}

\affiliation[1]{organization={Applied Artificial Intelligence Institute, Deakin University},
            addressline={75 Pigdons Rd},
            city={Waurn Ponds},
            postcode={3216},
            state={VIC},
            country={Australia}}


\makeatletter
\patchcmd{\ps@pprintTitle}
  {Preprint submitted to}
  {Preprint accepted at}
  {}{}
\makeatother
\begin{abstract}
Bayesian Optimization (BO) is an effective approach for the global optimization of black-box functions when function evaluations are expensive. Most prior works use Gaussian processes to model the black-box function, however, the use of kernels in Gaussian processes leads to two problems: first, the kernel-based methods scale poorly with the number of data points and second, kernel methods are usually not effective on complex structured high dimensional data due to curse of dimensionality. Therefore, we propose a novel black-box optimization algorithm where the black-box function is modeled using a neural network. Our algorithm does not need a \emph{Bayesian} neural network to estimate predictive uncertainty and is therefore computationally favorable. We analyze the theoretical behavior of our algorithm in terms of regret bound using advances in NTK theory showing its efficient convergence. We perform experiments with both synthetic and real-world optimization tasks and show that our algorithm is more sample efficient compared to existing methods.
\end{abstract}
\end{frontmatter}







\section{Introduction}
\label{sec1}
Optimizing a black-box function is crucial in many real-world application domains. Some well-known examples include hyper-parameter optimization of machine learning algorithms \cite{snoek2012practical,bergstra2012random}, synthesis of short polymer fiber materials, alloy design, 3D bio-printing, molecule design, etc \cite{greenhill2020bayesian, shahriari2015taking}. Bayesian Optimization (BO) has emerged as an effective approach to optimize such expensive-to-evaluate black-box functions. Most BO algorithms work as follows. First, a probabilistic regression model e.g. a Gaussian Process (GP) is trained on available function observations. This model is then used to construct an acquisition function that trades off between two potentially conflicting objectives: exploration and exploitation. The acquisition function is then optimized to suggest a point where the black-box functions should be next evaluated. There are several choices for acquisition functions, e.g., improvement-based schemes (Probability of Improvement \cite{kushner1964new}, Expected improvement \cite{mockus1978application}), Upper Confidence Bound scheme \cite{srinivas2009gaussian}, entropy search (ES \cite{hennig2012entropy}, PES \cite{wang2017max}), Thompson Sampling \cite{chowdhury2017kernelized} and Knowledge Gradient \cite{frazier2008knowledge}.
 
While GPs are a natural and usually the most prominent probabilistic model choice for BO, unfortunately, they are not suitable for optimizing functions in complex structured high dimensional spaces such as functions involving optimization of images and text documents. This is because GPs use kernels to model the functions and the popular kernels e.g. Square Exponential kernel, Mat\'ern kernel cannot accurately capture the similarity between complex structured data. This limitation prevents BO with GP from being applied to computer vision and natural language processing applications. Further, GPs do not scale well computationally due to kernel inversion requiring cubic complexity. 

Recently, Deep Neural Networks (DNNs) have emerged as state-of-the-art models for a wide range of tasks such as speech recognition, computer vision, natural language processing, and so on. The emergence of DNNs proves their power to scale and generalize to most kinds of data, including high dimensional and structured data as they can extract powerful features and scale linearly with the number of data points unlike GPs. Leveraging these advantages, some efforts have been made to utilize DNNs for optimization e.g., black-box function optimization in continuous search spaces \cite{snoek2015scalable, springenberg2016bayesian, pariagreedy} and contextual bandit optimizations in discrete search spaces \cite{zhou2020neural, xu2020neural, zhang2021neural, kassraie2022neural}. However, these works have several drawbacks: (a) 
some of these approaches are based on Bayesian Neural Networks (BNNs), which are computationally expensive as they typically require sampling methods to compute the posterior of networks' parameters. 
(b) other works in continuous search space settings do not provide any theoretical analysis of the respective optimization algorithms. While the first problem has been handled to some extent in \cite{springenberg2016bayesian,kim2021deep}, the second problem still remains open. Therefore, the problem of developing a DNN-based black-box optimization algorithm that is computationally efficient and comes with a mathematical guarantee of its convergence and sample efficiency remains open.


Addressing the above challenge, we propose a novel black-box optimization algorithm where the unknown function is modeled using an over-parameterized deep neural network. Our algorithm does not need a \emph{Bayesian Neural Network} to model predictive uncertainty and is therefore computationally favorable. We utilize the recent advances in neural network theory using Neural Tangent Kernels (NTK) to estimate a confidence interval around the neural network model and then use it for Thompson Sampling to draw a random sample of the function, which is maximized to recommend the next function evaluation point. We analyze the theoretical behavior of our algorithm in terms of its regret bound and show that our algorithm is convergent and more sample efficient compared to existing methods. In summary, our contributions are:

\begin{itemize}
    \item A deep neural network based black-box optimization (NeuralBO) that can perform efficient optimization by accurately modeling the complex structured data (e.g. image, text, etc) and also being computationally favorable, i.e. scaling linearly with the number of data points.
    \item A theoretical analysis of our proposed NeuralBO algorithm showing that our algorithm is convergent with a sub-linear regret bound. 
    \item Experiments with both synthetic benchmark optimization functions and real-world optimization tasks showing that our algorithm outperforms current state-of-the-art black-box optimization methods.
\end{itemize} 
\section{Related Works}
\subsection{Bayesian optimization}
Bayesian optimization is a well-established approach for performing the global optimization of noisy, expensive black-box functions  \cite{mockus1978application}. Traditionally, Bayesian optimization relies on the construction of a probabilistic surrogate model that places a prior on objective functions and updates the model using the existing function evaluations. To handle the promise and uncertainty of a new point, an acquisition function is used to capture the trade-off between exploration and exploitation through the posterior distribution of the functions. By performing a proxy optimization over this acquisition function, the next evaluation point is determined. The Expected Improvement (EI) \cite{mockus1978application} and Upper Confidence Bound (UCB) \cite{srinivas2009gaussian} criteria are typically used due to their analytical tractability in both value and gradient. Besides, the information-theoretic acquisition functions e.g., (ES \cite{hennig2012entropy}, PES \cite{hernandez2014predictive}, MES \cite{wang2017max}) guide our evaluations to locations where we can maximise our learning about the unknown minimum rather than to locations where we expect to obtain higher
function values. For a review on Bayesian optimization, see \cite{greenhill2020bayesian, shahriari2015taking,brochu2010tutorial, lizotte2008practical}.  
\subsection{Bayesian Optimization using GPs} 
Because of their flexibility, well-calibrated uncertainty, and analytical properties, GPs have been a popular choice to model the distribution over functions for Bayesian optimization. Formally, a function $f$ is assumed to be drawn from GP, i.e.,  $f \sim \text{GP}(m(\mathbf{x}), k(\mathbf{x}, \mathbf{x}^\prime))$, where $\mu(\mathbf{x})$ is a mean function and $k(\mathbf{x}, \mathbf{x}^\prime)$ is a covariance function. Common covariance functions include linear kernel, squared exponential (SE) kernel and Mat\'ern kernel. Thus, given a set of $t$ observed points $\mathcal{D}_{1:t} = \{\mathbf{x}_i, y_i\}_{i=1}^t$, the posterior mean and variance at step $t+1$ are computed as: $\mu_{t+1}(\mathbf{x}) = \mathbf{k}^\top \mathbf{K}^{-1} \mathbf{y}$ and variance $\sigma_{t+1}^{2}({\bf x}) = k({\bf x},{\bf x})-{\bf k}^{\top} \mathbf{K}^{-1}{\bf k}$, where $\mathbf{K} = \left[k(\mathbf{x}_i, \mathbf{x}_j)\right]_{1 \le i,j \le t}$, $\mathbf{k} = \begin{bmatrix}
k(\mathbf{x}, \mathbf{x}_1) & k(\mathbf{x}, \mathbf{x}_2) & \dots & k(\mathbf{x}, \mathbf{x}_t)
\end{bmatrix}$ and $\mathbf{y} = [y_1, y_2, \cdots, y_t]$.
Nonetheless, a major drawback of GP-based Bayesian optimization is that the computational complexity grows cubically in the number of observations, as it necessitates the inversion of the covariance matrix. Further, the covariance functions are usually too simplistic to model the function effectively when the inputs are high dimensional and complex structured data such as images and text.   
\subsection{Black-box Optimization using models other than GPs}
\label{section:BBO_without_GP}
There are relatively few works in black-box optimization that have used models other than GPs. For example, random forest (RF) and neural networks (NNs) have been used to estimate black-box functions. The first work that utilized RF as surrogate model was proposed in SMAC \cite{hutter2011sequential} and later refined in \cite{JMLR:v23:21-0888}.  Although random forest performs well by making good predictions in the neighbourhood of training data, it faces trouble with extrapolation \cite{shahriari2015taking}. Over the past decade, neural networks have emerged as a potential candidate to play the role of surrogate model in black-box optimization. The first work on black-box optimization using neural networks \cite{snoek2015scalable} utilizes a deep neural network as a mapping to transform high-dimensional inputs to lower-dimensional feature vectors before fitting a Bayesian linear regressor surrogate model. Another attempt has been made in \cite{springenberg2016bayesian} where the objective function has been modeled using a Bayesian neural network, which has high computational complexity due to the need to maintain an ensemble of models to get the posterior distribution over the parameters. Although these early works show the promise of neural networks for black-box optimization tasks through empirical results, they are still computationally unscalable (due to using Bayesian models) and no theoretical guarantees have been provided to ensure the convergence of their algorithms. 

Very recently, \cite{pariagreedy} also explores the use of deep neural networks as an alternative to GPs to model black-box functions. However, this work only provides theoretical guarantees for their algorithm under \emph{infinite width} of DNN setting, which is unrealistic in practice, while we provide theoretical guarantees for over-parameterized DNN, which is more general than their setting. Furthermore, \cite{pariagreedy} simply used a noise on the target does not correctly account for epistemic uncertainty and thus this method performs exploration randomly like  $\epsilon$-greedy, which is known to be sub-optimal. Moreover, their experimental results are limited to only synthetic functions, which is not enough to demonstrate the effectiveness of an algorithm in the real world. In contrast, we demonstrate the effectiveness of their algorithm on various real applications with high dimensional structured and complex inputs.   

In the field of deep learning, there is another line of research known as Neural Architecture Search (NAS) \cite{kandasamy2018neural, white2021bananas}. It is important to note that this line of research differs from the work here as NAS employs Bayesian Optimization or similar search/optimization algorithms for searching through a large space of potential neural network architectures to identify the optimal architecture that would be most suitable for a given dataset. The search process is guided by an objective function that typically represents the performance of a particular neural network architecture on a specific task or dataset. The objective of NAS is to automate the process of designing neural network architectures, and Bayesian Optimization has been shown to be an effective technique for achieving this goal. In contrast, our work here relates to using a deep neural network as a surrogate model in devising an effective Bayesian optimization algorithm.

\subsection{Contextual Bandits with Neural Networks}
There is a research line of contextual bandits which also uses deep neural networks to model reward functions. This problem is defined as follows: at each round $t$, the agent is presented with a set of $K$ actions associated with $K$ feature vectors called ``contexts''. After choosing an arm, the agent receives a scalar reward generated from some unknown function with respect to the corresponding chosen feature vector. The goal of the agent is to maximize the expected cumulative reward over $T$ rounds. For theoretical guarantees, most of the works in this line (e.g., \cite{zhou2020neural,zhang2021neural,xu2020neural,kassraie2022neural}) 
utilize a \emph{over-parametrized} DNN to predict rewards. While NeuralUCB \cite{zhou2020neural} and NeuralTS \cite{zhang2021neural} use UCB and Thompson Sampling acquisition functions to control the trade-off between exploration and exploitation by utilizing the gradient of the network in  algorithms, \cite{xu2020neural} performs LinearUCB on top of the representation of the last layer of the neural network, motivated by \cite{snoek2015scalable}. 

Similar to the above contextual bandit algorithms, we attempt to develop an effective algorithm with theoretical analysis for black-box optimization using over-parametrized neural networks with \emph{finite} width. The key challenge we need to address when extending the contextual bandit algorithms to black-box optimization is that of dealing with a continuous action space instead of a discrete space. Having continuous action space requires to bound the function modeling error at arbitrary locations in the action space. More specifically, from neural networks' viewpoint, this requirement is equivalent to characterizing predictive output on unseen data, which requires using neural network generalization theory different from the problem setting of contextual bandits.
\section{Preliminaries}
\subsection{Problem Setting}
\label{sec:problem_setting}
We consider a global optimization problem to maximize $f(\mathbf{x})$ subject to $\mathbf{x} \in \mathcal D \subset \mathbb{R}^d$, where $\mathbf{x}$ is the input with $d$ dimensions. The function $f$ is a noisy black-box function that can only be evaluated via noisy evaluations of $f$ in the form  $y= f(x) + \epsilon$, where $\epsilon$ is a sub-Gaussian noise which we will discuss more clearly in Assumption 2 in our regret analysis section. In our work, we consider input space $\mathcal D$ of the form: $a \leq \norm{\mathbf{x}}_2 \leq  b$,  where $a, b$ are absolute constants and $a \le b$. We note that this assumption is mild compared to current works in neural contextual bandits \cite{zhou2020neural,zhang2021neural,xu2020neural,kassraie2022neural} that required that $\norm{\mathbf{x}}_2 = 1$.  

To imply the smoothness of function $f$, we assume the objective function $f$ is from an RKHS corresponding to a positive definite Neural Tangent Kernel $k_\text{NTK}$. In particular, $\norm{f}_{\mathcal{H}_{k_\text{NTK}}} \leq B$, for a finite $B > 0$. These assumptions are regular and have been typically used in many previous works \cite{chowdhury2017kernelized, srinivas2009gaussian,vakili2021information}.

\subsection{RKHS with Neural Tangent Kernel}
\label{subsection:NTK}

Given a training set $\mathcal{X}_t = \{(\mathbf{x}_i, y_i)\}^t_{i=1} \subset \mathcal{D} \times \mathbb{R}$, that is a set of $t$ arbitrary points picked from $\mathcal D$ with its corresponding observations, and consider training a neural network $h(\mathbf{x}; \boldsymbol{\theta})$ with gradient descent using an infinitesimally small learning rate. The Neural Tangent Kernel (NTK) $k_{\text{NTK}}(\mathbf{x}, \mathbf{x}^\prime)$ is defined as:  
$k_{\text{NTK}}(\mathbf{x}, \mathbf{x}^\prime)= \langle \mathbf{g}(\mathbf{x}; \boldsymbol{\theta}), \mathbf{g}(\mathbf{x}^\prime; \boldsymbol{\theta}) \rangle/m$, where $\mathbf{g}(\mathbf{x}; \boldsymbol{\theta})$ is the gradient of $h(\mathbf{x}; \boldsymbol{\theta})$ as defined. When the network width $m$ increases,  $\langle \mathbf{g}(\mathbf{x}; \boldsymbol{\theta}), \mathbf{g}(\mathbf{x}^\prime, \boldsymbol{\theta}) \rangle$ gradually converges to $\langle \mathbf{g}(\mathbf{x}, \boldsymbol{\theta}_0), \mathbf{g}(\mathbf{x}^\prime, \boldsymbol{\theta}_0) \rangle$
and the corresponding NTK matrix is defined in Definition \ref{def:NTK_matrix}.

\begin{definition} [\cite{jacot2018neural}]
\label{def:NTK_matrix}
Consider the same set $\mathcal{X}_t$ as in Section \ref{subsection:NTK} . Define

\[ \widetilde{\mathbf{H}}_{i,j}^{(1)} = \mathbf{\Sigma}_{i,j}^{(1)} = \langle  \mathbf{x}_i, \mathbf{x}_j \rangle\ , \mathbf{A}_{i,j}^{(l)} = 
\begin{pmatrix}
\mathbf{\Sigma}_{i,i}^{(l)} & \mathbf{\Sigma}_{i,j}^{(l)} 
\\
\mathbf{\Sigma}_{i,j}^{(l)} & \mathbf{\Sigma}_{j,j}^{(l)}
\end{pmatrix}\] 

\[\mathbf{\Sigma}_{i,j}^{(l+1)} = 2 \mathbb{E}_{(u,v) \sim \mathbf{N}(\mathbf{0}, \mathbf{A}_{i,j}^{(l)})} [\phi(u) \phi(v)] \]

\[ \widetilde{\mathbf{H}}_{i,j}^{(l+1)} = 2\widetilde{\mathbf{H}}_{i,j}^{(l)}\mathbb{E}_{(u,v) \sim \mathbf{N}(\mathbf{0}, \mathbf{A}_{i,j}^{(l)}} [\phi^\prime(u) \phi^\prime(v)] + \mathbf{\Sigma}_{i,j}^{(l+1)}, \]
where $\phi(\cdot)$ is the activation function of the neural network $h(\mathbf{x}; \boldsymbol{\theta})$.
Then $\mathbf{H}_t = \frac{\mathbf{\widetilde{H}}^{(L)}+ \mathbf{\Sigma} ^ {(L)}}{2}$ is called the NTK matrix on the set $\mathbf{X}$. 
\end{definition}

We assume $f$ to be a member of the Reproducing Kernel Hilbert Space (RKHS) of real-valued functions on $\mathcal{D}$ with Neural Tangent Kernel (NTK, see Section \ref{subsection:NTK}), and a bounded norm, $\norm{f}_{\mathcal{H}_{k_\textup{NTK}}} \leq B$.  This RKHS, denoted by $\mathcal{H}_{k_\textup{NTK}}(\mathcal D)$, is completely specified by its kernel function $k_\textup{NTK}(\cdot, \cdot)$ and vice-versa, with an inner product $\langle \cdot, \cdot \rangle$ obeying the reproducing property:
$f(\mathbf{x}) = \langle f, k_\textup{NTK}(\cdot, \mathbf{x})\rangle$ for all $f \in  \mathcal{H}_{k_\textup{NTK}}(\mathcal{D})$. Equivalently, the RKHS associated with $k_\textup{NTK}$ is then given by:
\[ \mathcal{H}_{k_\textup{NTK}} = \{f: f = \sum_i \alpha_i k_{\textup{NTK}}(\mathbf{x}_i, \cdot), \alpha_i \in \mathbb{R}\}\]
The induced
RKHS norm $\norm{f}_{\mathcal{H}_{k_\textup{NTK}}} = \sqrt{\langle f,f\rangle_{\mathcal{H}_{k_\textup{NTK}}}}$ measures the smoothness of $f$, with respect to the kernel function $k_\textup{NTK}$, and satisfies: $f \in \mathcal{H}_{k_\textup{NTK}}(\mathcal{D})$ if and only if $\norm{f}_{k_{\mathcal{H}_\textup{NTK}}} < \infty$. 
\subsection{Maximum Information Gain}
Assume after $t$ steps, the model receives an input sequence $\mathcal{X}_t = (\mathbf{x}_1, \mathbf{x}_2, \dots  \mathbf{x}_t)$ and observes noisy rewards $\mathbf{y}_t = (y_1, y_2, \dots, y_t)$. The \emph{information gain} at step $t$, quantifies the reduction in uncertainty about $f$ after observing $\mathbf{y}_t$, defined as the mutual information between  $\mathbf{y}_t$ and $f$:
\[
I(\mathbf{y}_t; f):=  H(\mathbf{y}_t) - H(\mathbf{y}_t \rvert f), 
\]
where $H$ denotes the entropy. To obtain the closed-form expression of information gain, one needs to introduce a GP model where $f$ is assumed to be a zero mean GP indexed on $\mathcal X$ with kernel $k$. Let $\mathbf{f}_t = \left(f(\mathbf{x}_1), f(\mathbf{x}_1), \dots, f(\mathbf{x}_t)] \right)$ be the corresponding true function values.  From \cite{cover1999elements}, the mutual information between two multivariate Gaussian distributions is: 
\[ I(\mathbf{y}_t; f) = I(\mathbf{y}_t; \mathbf{f}_t) = \frac{1}{2} \log \det (\mathbf{I} + \lambda^{-1}\mathbf{H}_t), \]
where $\lambda > 0$ is a regularization parameter and $\mathbf{H}_t$ is the kernel matrix. In our case, $\mathbf{H}_t$ can be referred to as the NTK matrix associated with Neural Tangent Kernel.

Assuming GP model with NTK kernel to calculate closed-form expression of maximum information gain is popular in NTK-related works \cite{kassraie2022neural, kassraie2022graph}. 

To further achieve  input-independent and kernel-specific bounds, we define the \emph{maximum information gain} as follows: The \emph{maximum information gain} $\gamma_t$, after $t$ steps, defined as: 
\[ \gamma_t := \max_{\mathcal{A} \subset \mathcal{X}, \lvert A \rvert = t} I(\mathbf{y}_A; \mathbf{f}_A)\]

As stated in Lemma 3 of \cite{chowdhury2017kernelized}, the maximum information gain $\gamma_t$ can be lower bounded as:
\[\gamma_t \ge \frac{1}{2} \log \det (\mathbf{I} + \lambda^{-1} \mathbf{H}_t) \]


\section{Proposed NeuralBO Algorithm}
\label{section:algorithm}

\begin{algorithm}[]
\caption{Neural Black-box Optimization (NeuralBO)}
\label{alg:NeuralBO}
\textbf{Input}: The input space $\mathcal D$, the number of rounds $T$, exploration parameter $\nu_t$, network width $m$, RKHS norm bound $B$, regularization parameter $\lambda$, $\alpha \in (0,1)$.
\begin{algorithmic}[1]
\State Set $\mathbf{U}_0 = \lambda \mathbf{I}$ 

\For{$t = 1$ to $T$}
\State $\sigma^2_t (\mathbf{x}) = \lambda \mathbf{g}(\mathbf{x};\boldsymbol{\theta}_0)^\top\mathbf{U}_{t-1}^{-1}\mathbf{g}(\mathbf{x};\boldsymbol{\theta}_0)/m$ \label{line:calculate_var}

\State $\widetilde{f_t}(\mathbf{x}) \sim \mathcal{N}(h(\mathbf{x}; \boldsymbol{\theta}_{t-1}), \nu_t^2 \sigma^2_t(\mathbf{x}))$ 
\State Choose $\mathbf{x}_t = \argmax_{\mathbf{x} \in \mathcal{D}} \widetilde{f_t}(\mathbf{x}) $ and receive observation $y_t = f(\mathbf{x}_t) + \epsilon_t$ 
   
\State Update $\boldsymbol{\theta}_t \leftarrow \text{TrainNN}(\{\mathbf{x}_i\}^t_{i=1} \{y_i\}^t_{i=1}, \boldsymbol{\theta}_0)$ \label{line:train_NN}
\State $\mathbf{U}_t \leftarrow \mathbf{U}_{t-1} +  \frac{\mathbf{g}(\mathbf{x}_t;\boldsymbol{\theta}_0)\mathbf{g}(\mathbf{x}_t; \boldsymbol{\theta}_0)^\top}{m} $ \label{line:update_Ut}
\EndFor
\end{algorithmic}
\end{algorithm}

\begin{algorithm}[H]
\caption{TrainNN}
\label{alg:train_NN}
\textbf{Input}: Chosen points $\{\mathbf{x}_i\}^t_{i=1}$, noisy observations $\{y_i\}^t_{i=1}$, initial parameter $\boldsymbol{\theta}_0$, number of gradient descent steps $J$, regularization parameter $\lambda$, network width $m$, learning rate $\eta$.
\begin{algorithmic}[1]
\State Define objective function $\mathcal{L}(\boldsymbol{\theta})  = \frac{1}{2}\sum^T_{i=1} (f(\mathbf{x}_i; \boldsymbol{\theta}) - y_i)^2 + \frac{1}{2}m\lambda\norm{\boldsymbol{\theta} - \boldsymbol{\theta}^{(0)}}^2_2$
\For{$k=0,\cdots ,J-1$}

$\boldsymbol{\theta}^{(k+1)} = \boldsymbol{\theta}^{(k)} - \eta \nabla \mathcal{L}(\boldsymbol{\theta}^{(k)})$
\EndFor

\State return $\boldsymbol{\theta}^{(J)}$.
\end{algorithmic}
\end{algorithm}

In this section, we present our proposed neural-network based black-box algorithm: NeuralBO. Following the principle of BO, our algorithm consists of two main steps: (1) Build a model of the black-box objective function, and (2) Use the black-box model to choose the next function evaluation point at each iteration. For the first step, unlike traditional BO algorithms which use Gaussian process to model the objective function, our idea is to use a fully connected neural network $h(\mathbf{x}; \boldsymbol{\theta})$ to learn the function $f$ as follows:
$$ h(\mathbf{x};\boldsymbol{\theta}) = \sqrt{m} \mathbf{W}_L \phi(\mathbf{W}_{L-1}\phi(\cdots\phi(\mathbf{W}_1 \mathbf{x})), $$
where $\phi(x) = \max(x,0)$ is the rectified linear unit (ReLU) activation function, $\mathbf{W}_1 \in \mathbb{R}^{m \times d}, \mathbf{W}_i \in \mathbb{R}^{m \times m}, 2\leq i \leq L-1, \mathbf{W}_L \in \mathbb{R}^{1 \times m}$, and $\boldsymbol{\theta} = (vec(\mathbf{W}_1),\cdots, vec(\mathbf{W}_L)) \in \mathbb{R}^p$ is the collection of parameters of the neural network, $p=md+m^2(L-2)+m$. To keep the analysis convenient, we assume that the width $m$ is the same for all hidden layers. We also denote the gradient of the neural network by $\mathbf{g}(\mathbf{x}; \boldsymbol{\theta}) = \nabla_{\boldsymbol{\theta}}h(\mathbf{x}; \boldsymbol{\theta}) \in \mathbb{R}^p$.

For the second step, we choose the next sample point $\mathbf{x}_t$ by using a Thompson Sampling strategy. In particular, given each $\mathbf{x}$, our algorithm maintains a Gaussian distribution for $f(\mathbf{x})$. To choose the next point, it samples a random function $\widetilde{f}_t$ from the posterior distribution $\mathcal{N}(h(\mathbf{x}, \boldsymbol{\theta}_{t-1}), \nu_t^2 \sigma^2_t (\mathbf{x}))$, where $\sigma^2_t (\mathbf{x}) = \lambda \mathbf{g}(\mathbf{x};\boldsymbol{\theta}_0)\mathbf{U}_{t-1}^{-1}\mathbf{g}(\mathbf{x};\boldsymbol{\theta}_0)/m$ and $\nu_t = \sqrt{2}B +\frac{R}{\sqrt{\lambda}}( \sqrt{2 \log(1/ \alpha)}$. From here, $\sigma_t(\mathbf{x})$ and $\nu_t$ construct a confidence interval, where for all $\mathbf{x} \in \mathcal{D}$, we have $\lvert  f(\mathbf{x}) - f(\mathbf{x}, \boldsymbol{\theta}) \rvert \leq \nu_t\sigma_t(\mathbf{x})$ that holds with probability $1-\alpha$. We also note the difference from \cite{zhang2021neural} that uses $\mathbf{g}(\mathbf{x};\boldsymbol{\theta}_t)/\sqrt{m}$ as dynamic feature map, we here use a fixed feature map $\mathbf{g}(\mathbf{x};\boldsymbol{\theta}_0)/\sqrt{m}$ which can be viewed as a finite approximation of the feature map $\phi(\mathbf{x})$ of $k_\textup{NTK}$. This allows us to eliminate $\sqrt{m}$ away the regret bound of our algorithm. Our algorithm chooses $\mathbf{x}_t$ as $\mathbf{x}_t = \argmax_{\mathbf{x} \in \mathcal{D}} \widetilde{f_t}(\mathbf{x})$. 


Besides, to initialize the network $h(\mathbf{x}; \boldsymbol{\theta}_0)$, we randomly generate each element of $\boldsymbol{\theta}_0$ from an appropriate Gaussian distribution: for each $1 \leq l \leq L-1$, each entry of $\mathbf{W}$ is generated independently from $\mathcal N(0, 2/m)$; while each entry of the last layer $\mathbf{W}_L$ is set to zero to make $h(\mathbf{x}, \boldsymbol{\theta}_0) = 0$. Here we inherit so-called \emph{He initialization} \cite{he2015delving}, which ensures the stability of the expected length of the output vector at each hidden layer. 

The proposed NeuralBO is summarized using Algorithm \ref{alg:NeuralBO} and Algorithm \ref{alg:train_NN}. In section \ref{sec:regret_analysis}, we will demonstrate that our algorithm can achieve a sub-linear regret bound, and  works well in practice (See our section \ref{section:experiments}).
\paragraph{Discussion}
A simple way to have a neural network based black-box algorithm is to extend the existing works of \cite{zhou2020neural,zhang2021neural} or \cite{kassraie2022neural} to our setting where the search space is continuous. However, this is difficult because these algorithms are designed for a finite set of actions. A simple adaptation can yield non-vanishing bounds on cumulative regret. For example, the regret bound in \cite{kassraie2022neural} depends on the size of finite action spaces $\lvert \mathcal A \rvert$. If $\lvert \mathcal A \rvert \rightarrow \infty$ then the regret bound goes to infinity. In \cite{zhou2020neural,zhang2021neural}, the regret bounds are built on gram matrix $\mathbf{H}$ which is defined through the set of actions. However, such a matrix cannot even be defined in our setting where the search space (action space) is infinite. We solve this challenge by using a discretization of the search space at each iteration as mentioned in Section \ref{sec:regret_analysis}. In addition, for our confidence bound calculation, using $\mathbf{g}(\mathbf{x};\boldsymbol{\theta}_0)/\sqrt{m}$ instead of $\mathbf{g}(\mathbf{x};\boldsymbol{\theta}_t)/\sqrt{m}$ as in \cite{zhou2020neural,zhang2021neural} allows us to reduce a factor $\sqrt{m}$ from our regret bound.    

Our algorithm is also different from \cite{chowdhury2017kernelized} in traditional Bayesian optimization which uses a Gaussian process to model the function with posterior computed in closed form. In contrast, our algorithm computes posterior by gradient descent. 





\section{Regret Analysis}
\label{sec:regret_analysis}
In this section, we provide a regret bound for the proposed NeuralBO algorithm. Our regret analysis is built upon the recent advances in NTK theory \cite{allen2019convergence, cao2019generalization} and proof techniques of GP-TS \cite{chowdhury2017kernelized}. 
Unlike most previous neural-based contextual bandits works \cite{zhang2021neural, zhou2020neural} that restrict necessarily the search space to a set $\mathbb{S}^{d-1} = \{ \mathbf{x} \in \mathbb {R}^d \colon \norm{\mathbf{x}}_2 = 1 \}$, we here derive our results for a flexible condition, where the norm of inputs are bounded: $a \leq \norm{\mathbf{x}}_2 \leq b, \text{where } 0 < a \le b$.


For theoretical guarantees, we need to use the following assumptions on the observations $\{\mathbf{x}_i\}^T_{i=1}$.
\begin{assumption} \label{assumption:sufficient_exploration} There exists $\lambda_0 > 0$, 
$\mathbf{H}_T \ge \lambda_0 \mathbf{I}_T$.
\end{assumption}
This is a common assumption in the literature \cite{zhou2020neural,xu2020neural,zhang2021neural}, and is satisfied as long as we sufficiently explore the input space such that no two inputs $\mathbf{x}$ and $\mathbf{x}^\prime$ are identical.
\begin{assumption}
\label{assumption:iid_noise}
We assume the noises $\{\epsilon_i\}_{i=1}^T$ where $\epsilon_i = y_i - f(\mathbf{x}_i)$, are i.i.d and sub-Gaussian with parameter $R > 0$ and are independent with $\{\mathbf{x}_i\}_{i=1}^T$. 
\end{assumption} 
This assumption is mild and similar to the work of \cite{vakili2021optimal} where the assumption is used to prove an optimal order of the regret bound for Gaussian process bandits. We use it here to provide a regret bound with the sublinear order for our proposed algorithm.    

Now we are ready to bound the regret of our proposed NeuralBO algorithm. To measure the regret of the algorithm, we use the cumulative regret which is defined as follows: $R_T = \sum_{t=1}^T r_t $ after $T$ iterations, where $\mathbf{x}^* = \argmax_{\mathbf{x} \in \mathcal{D}} f(\mathbf{x}) $ is the maximum point of the unknown function $f$ and $r_t = f(\mathbf{x^*}) - f(\mathbf{x}_t)$ is instantaneous regret incurred at time $t$. We present our main theoretical result.
\newtheorem{thm}{Theorem}
\begin{thm}
\label{theorem:main} Let $\alpha \in (0,1)$. Assume that the true underlying $f$ lies in the RKHS $\mathcal{H}_{k_\textup{NTK}}(\mathcal D)$ corresponding
to the NTK kernel $k_{\textup{NTK}}(\mathbf{x}, \mathbf{x'})$ with RKHS norm bounded by $B$.
Set the parameters in Algorithm \ref{alg:NeuralBO}
as $\lambda = 1 + \frac{1}{T}$, $\nu_t = \sqrt{2}B +\frac{R}{\sqrt{\lambda}}( \sqrt{2 \log(1/ \alpha)}$, where $R$ is the noise sub-Gaussianity parameter.  
Set $\eta = (m\lambda + mLT)^{-1}$ and $J = (1+ LT /\lambda) \left(1+ \log(T^3L\lambda^{-1}\log \left(1/ \alpha \right))\right)$. If the network width m satisfies:
\[ m \geq \textup{poly}\left(\lambda, T, \log(1/\alpha), \lambda_0^{-1}\right),\]
then with probability at least $1-\alpha$, the regret of Algorithm \ref{alg:NeuralBO} is bounded as

\begin{flalign*}
R_T & \leq C_1(1+c_T)\nu_T \sqrt{L} \sqrt{\frac{\lambda BT}{\log(B+1)} (2\gamma_T+1)}  \\
     & +  (4+  C_2(1+c_T)\nu_T L)\sqrt{2 \log(1/\alpha)T} + \frac{(2B+1)\pi^2}{6} + \frac{b(a^2+b^2)}{a^3}
\end{flalign*} 
where $C_1, C_2$ are absolute constants, $a$ and $b$ are lower and upper norm bounds of $\mathbf{x} \in \mathcal{D}$  and $c_T=\sqrt{4\log T + 2 \log \ \lvert \mathcal D_t \rvert}$.
\end{thm}
\remark{There exists a component $\frac{b(a^2 + b^2)}{a^3}$ in the regret bound $R_T$. This component comes from the condition of the search space. If we remove the influence
of constants $a,b, B$ and $R$, Theorem \ref{theorem:main} implies that our regret bound follows an order $\widetilde{O}( \sqrt{T\gamma_T})$. This regret bound has the same order as the one of the MVR algorithm in \cite{vakili2021optimal}, but is significantly better than traditional BO algorithms (e.g., \cite{srinivas2009gaussian, chowdhury2017kernelized}) 
 and existing neural contextual bandits (e.g., \cite{zhou2020neural,zhang2021neural}) that have the order $\widetilde{O}(\gamma_T\sqrt{T})$.  We note that the regret bounds in \cite{zhou2020neural,zhang2021neural} are expressed through the effective dimension $\widetilde{d} = \frac{ \log \det (\mathbf{I} + \mathbf{H}/\lambda)}{\log(1 + TK/\lambda)}$, where $K$ is the number of arm in contextual bandits setting. However, $\gamma_T$ and the effective dimension are of the same order up to a ratio of $1/(2 \text{log} (1 + TK/\lambda))$. The improvement of factor $\sqrt{\gamma_T}$ is significant because for NTK kernel as shown by \cite{kassraie2022neural}, $\gamma_T$ follows order $\mathcal{O}(T^{1-1/d})$, where $d$ is the input dimension. For $d \ge 2$, existing bounds become vacuous and are no longer sub-linear. In contrast, our regret bound remains sub-linear for any $d>0$. 

} 
\section{Proof of the Main Theorem}
To perform analysis for continuous actions as in our setting, we use a discretization technique. At each time $t$, we use a discretization $\mathcal{D}_t \subset \mathcal{D}$ with the property that $\lvert f(\mathbf{x}) - f([\mathbf{x}]_t) \rvert \leq \frac{1}{t^2}$ where $[\mathbf{x}]_t \in \mathcal{D}_t$ is the closest point to $\mathbf{x} \in \mathcal{D}$. Then we choose $\mathcal{D}_t $ 
with size $\lvert \mathcal{D}_t \rvert = \left( (b-a)BC_{\text{lip}}t^2 \right)^d$ that satisfies $\norm{\mathbf{x} - [\mathbf{x}]_t}_1 \leq \frac{b-a}{(b-a)BC_{\text{lip}}t^2} = \frac{1}{BC_{\text{lip}}t^2}$ for all $\mathbf{x} \in \mathcal{D}$,
where $C_{\text{lip}}= \underset{\mathbf{x} \in \mathcal{D}} {\supremum} \underset{j \in [d]} {\supremum} \left( \frac{\partial^2 k_\text{NTK}(\mathbf{p},\mathbf{q})}{\partial \mathbf{p}_j \mathbf{q}_j} \lvert \mathbf{p}=\mathbf{q}=\mathbf{x} \right)$. This implies, for all $\mathbf{x} \in \mathcal{D}$:
\begin{equation}
\label{eqn:rkhs_lipschitz}
    \lvert f(\mathbf{x}) - f([\mathbf{x}]_t) \rvert \leq \norm{f}_{\mathcal{H}_{k_\textup{NTK}}} C_{\text{lip}} \norm{\mathbf{x} - [\mathbf{x}]_t}_1 \leq BC_{\text{lip}} \frac{1}{BC_{\text{lip}}t^2} =  1/t^2, 
\end{equation}
is Lipschitz continuous of any $f \in \mathcal{H}_{k_\textup{NTK}}(\mathcal{D})$ with Lipschitz constant $B C_{\text{lip}}$, where we have used the inequality $\norm{f}_{\mathcal{H}_{k_\textup{NTK}}} \leq B $  which is our assumption about function $f$. We bound the regret of the proposed algorithm by starting from the instantaneous regret $r_t = f(\mathbf{x}^*) - f(\mathbf{x}_t)$ can be decomposed to $r_t = [f(\mathbf{x}^*) - f([\mathbf{x}^*]_t)] + [f([\mathbf{x}^*]_t - f(\mathbf{x}_t)]$. While the first term is bounded by Eq(\ref{eqn:rkhs_lipschitz}),  we bound the second term in Lemma \ref{lemma:regret_bound} provided in our Appendix. The proof of Lemma \ref{lemma:regret_bound} requires a few steps - we introduce a saturated set $\mathcal S_t$ (see Definition \ref{def:saturated_set}), we then combine it with the results of Lemma \ref{lemma:sampling_bound} (deriving a bound on $\lvert \widetilde{f}_t(\mathbf{x}) - h(\mathbf{x}; \boldsymbol{\theta}_{t-1}) \rvert$) necessitated due to the Thompson sampling and Lemma \ref{lemma:predictive_bound} (deriving a bound on $\lvert f(\mathbf{x}) - h(\mathbf{x}; \boldsymbol{\theta}_{t-1}) \rvert$) utilizing the generalization result of the over-parameterized neural networks. It is noted that in our proof, we consider the effect of our general search space setting (where input $\mathbf{x} \in \mathcal{D}$ is norm-bounded, i.e.,  $0 < a \le \norm{\mathbf{x}}_2 \le b$) on the output of the over-parameterized deep neural network at each layer in Lemma \ref{lemma:bound_hil} and on the gradient of this network utilized in Lemma \ref{lemma:NN_vs_linear}. These results contribute to the appearance of the lower and the upper norm bounds $a$ and $b$, the first in the generalization property of over-parameterized neural networks in Lemma \ref{lemma:predictive_bound} and the latter affect the cumulative regret bound in Theorem \ref{theorem:main}. Our proof follows the proof style of Lemma 4.1 of \cite{cao2019generalization}, Lemma B.3 of \cite{cao2019generalization} and Theorem 5 
of \cite{allen2019convergence}. 

On the other hand, Lemma \ref{lemma:noise_affeted_bound}, based on Assumption \ref{assumption:iid_noise}, provides a tighter bound on the confidence interval that eliminated the factor $\sqrt{\gamma_T}$, in comparison with previous relevant works \cite{chowdhury2017kernelized, zhou2020neural}, leads to the sub-linear regret bound in Theorem \ref{theorem:main}.   
By these arguments, we achieve a cumulative regret bound for our proposed algorithm in terms of the maximum information gain $\gamma_T$ (Lemma \ref{lemma:regret_expectation}, \ref{lemma:regret_bound}, \ref{lemma:min_sigma}). \textbf{Our detailed proof is provided in \ref{sec:main_appendix}}.


\newproof{pthm}{Proof Sketch for Theorem \ref{theorem:main}}

\begin{pthm}
\label{proof:theorem_main}
With probability at least $1-\alpha$, we have
\begin{align*}
 R_T &  = \sum^T_{t=1} f(\mathbf{x^*}) - f(\mathbf{x}_t) \\ 
     & = \sum^T_{t=1} \left[f(\mathbf{x}^*) - f([\mathbf{x}^*]_t)\right] + \left[f([\mathbf{x}^*]_t) - f(\mathbf{x}_t) \right] \\ 
     &\leq 4T \epsilon(m) + \frac{(2B+1)\pi^2}{6} + \Bar{C_1}(1+c_T)\nu_T \sqrt{L} \sum^T_{i=1} \min(\sigma_t(\mathbf{x}_t), B) \\
     & \quad +(4+\Bar{C_2}(1+c_T)\nu_T L + 4\epsilon(m))\sqrt{2 \log(1/\alpha)T} \\
     & \leq \Bar{C_1}(1+c_T)\nu_T \sqrt{L} \sqrt{\frac{\lambda BT}{\log(B+1)} (2\gamma_T+1)} 
     + \frac{(2B+1)\pi^2}{6} + 4T \epsilon(m) \\
     & \quad + 4\epsilon(m)\sqrt{2 \log(1/\alpha)T}  +  \left(4+\Bar{C_2}(1+c_T)\nu_T L\right)\sqrt{2 \log(1/\alpha)T}  \\
     &  = \Bar{C_1}(1+c_T)\nu_t \sqrt{L} \sqrt{\frac{\lambda BT}{\log(B+1)} (2\gamma_T+1)} + \frac{(2B+1)\pi^2}{6} \\
     & \quad +  \epsilon(m)(4T+ \sqrt{2 \log(1/\alpha)T}) + (4+\Bar{C_2}(1+c_T)\nu_t L)\sqrt{2 \log(1/\alpha)T} \\
     \end{align*} 
The first inequality is due to Lemma \ref{lemma:regret_bound}, which provides the bound for cumulative regret $R_T$ in terms of $\sum^T_{t=1} \min(\sigma_t(\mathbf{x}_t),B)$.  The second inequality further provides the bound of term $\sum^T_{t=1} \min(\sigma_t(\mathbf{x}_t),B)$ due to Lemma \ref{lemma:min_sigma}, while the last equality rearranges addition.   Picking $\eta = (m\lambda + mLT)^{-1}$ and $J = \left(1+LT/\lambda \right) \left(\log (C_{\epsilon,2} ) + \log(T^3L\lambda^{-1}\log(1/\alpha)) \right)$, we have 
\begin{equation*}
\begin{split}
      &\frac{b}{a} C_{\epsilon,2}(1 - \eta m \lambda)^J \sqrt{TL/\lambda} \left(4T+\sqrt{2 \log(1/\alpha)T}\right)\\
    = & \frac{b}{a} C_{\epsilon,2} \left(1-\frac{1}{1+LT/\lambda}\right)^{J} \left(4T+\sqrt{2 \log(1/\alpha)T}\right) \\
    = & \frac{b}{a} C_{\epsilon,2} e^{-\left(\log \left(C_{\epsilon,2}\right) + \log(T^3L\lambda^{-1}\log(1/\alpha)) \right)} \left(4T+\sqrt{2 \log(1/\alpha)T}\right)\\
    = & \frac{b}{a}  \frac{1}{C_{\epsilon,2}}.T^{-3}L^{-1}\lambda \log^{-1}(1/\alpha) \left(4T+\sqrt{2 \log(1/\alpha)T}\right)  \le   \frac{b}{a}\\
\end{split}
\end{equation*}
Then choosing $m$ that satisfies:
\begin{equation*}
    \begin{split}
        \left(\frac{b}{a} C_{\epsilon,1} m^{-1/6}\lambda^{-2/3}L^3 \sqrt{\log m} + \left(\frac{b}{a}\right)^3 C_{\epsilon,3} m^{-1/6} \sqrt{\log m} L^4 T^{5/3} \lambda^{-5/3} (1+\sqrt{T/\lambda})\right) \\
        \left(4T+  \sqrt{2 \log(1/\alpha)T}\right) \le \left(\frac{b}{a}\right)^3 
    \end{split}
\end{equation*}
We finally achieve the bound of $R_T$ as:
\begin{flalign*}
R_T & \leq \Bar C_1(1+c_T)\nu_T \sqrt{L} \sqrt{\frac{\lambda BT}{\log(B+1)} (2\gamma_T+1)}  \\
     & +  (4+ \bar C_2(1+c_T)\nu_T L)\sqrt{2 \log(1/\alpha)T} + \frac{(2B+1)\pi^2}{6} + \frac{b(a^2+b^2)}{a^3}
\end{flalign*}
\end{pthm}

\section{Experiments}
\label{section:experiments}

In this section, we demonstrate the effectiveness of our proposed NeuralBO algorithm compared to traditional BO algorithms and several other BO algorithms on various synthetic benchmark optimization functions and various real-world optimization problems that have complex and structured inputs. The real-world optimization problems consist of (1) generating sensitive samples to detect tampered model trained on MNIST \cite{lecun-mnisthandwrittendigit-2010} dataset; (2) optimizing control parameters for robot pushing considered in \cite{wang2017max}; and (3) optimizing the number of evaluations needed to find a document that resembles an intended (unknown) target document.

For all experiments, we compared our algorithm with common classes of surrogate models used in black-box optimization, including Gaussian Processes (GPs), Random Forests and Deep Neural Networks (DNNs). For GPs, we have three baselines: GP-UCB \cite{srinivas2009gaussian}, GP-TS \cite{chowdhury2017kernelized} and GP-EI \cite{jones1998efficient} with the common Squared Exponential (SE) Kernel. Our implementations for GP-based Bayesian Optimization baselines utilize public library GPyTorch  \footnote{\url{https://gpytorch.ai/}} and BOTorch \footnote{\url{https://botorch.org/}}. Next, we consider SMAC \cite{hutter2011sequential} as a baseline using RF as a surrogate model. Lastly, we also include two recent DNNs-based works for black-box optimization: DNGO \cite{snoek2015scalable} and Neural Greedy \cite{pariagreedy}. We further describe the implementations for RF and DNN-based baselines as follows: 
\begin{itemize}
    \item DNGO \cite{snoek2015scalable} models the black-box function by a Bayesian linear regressor, using low-dimensional features (extracted by DNN) as inputs. We run DNGO algorithm with the implementation of AutoML \footnote{\url{https://github.com/automl/pybnn}} with default settings. 
    \item NeuralGreedy \cite{pariagreedy} fits a neural network to the current set of observations, where the function values are randomly perturbed before learning the neural network. The learnt neural network is then used as the acquisition function to determine the next query point. Our re-implementation of NeuralGreedy follows the setting described in Appendix F.2 \cite{pariagreedy}.  
    \item Random forest (RF) based surrogate model is implemented using public library GPyOpt \footnote{\url{https://github.com/SheffieldML/GPyOpt}} and optimization is performed with EI acquisition function.   
\end{itemize}

For our proposed NeuralBO algorithm, we model the functions using two-layer neural networks $f(\mathbf{x}; \boldsymbol{\theta}) = \sqrt{m}\mathbf{W}_2\sigma(\mathbf{W}_1\mathbf{x})$ with network width $m=500$. 
The weights of the network are initialized with independent samples from a normal distribution $\mathcal{N} (0, 1/m)$. To train the surrogate neural network models, we use SGD optimizer with batch size 50, epochs 50, learning rate $\eta=0.001$ and regularizer $\lambda=0.01$. We set exploration parameter in Algorithm \ref{alg:NeuralBO}: $\nu_t = \nu$ and do a grid search over $\{0.1,1,10\}$. 



\subsection{Synthetic Benchmark Functions}
\label{exp:optimize_syn_funcs}
We optimized several commonly used synthetic benchmark functions: Ackley, Levy, and Michalewics. The exact expression of these functions can be found at: \url{http://www.sfu.ca/~ssurjano/optimization.html}. For each function, we perform optimization in two different dimensions. The latter dimension is twice as high as the former. This is done to evaluate the methods across a range of dimensions. To emphasize the adaptation of our method to various dimensions, for each function, we performed optimization for four different dimensions, including 10, 20, 50, and 100.
The function evaluation noise follows a normal distribution with zero mean and the variance is set to 1\% of the function range.
\begin{figure}[H]
    \centering
    \includegraphics[width=\textwidth]{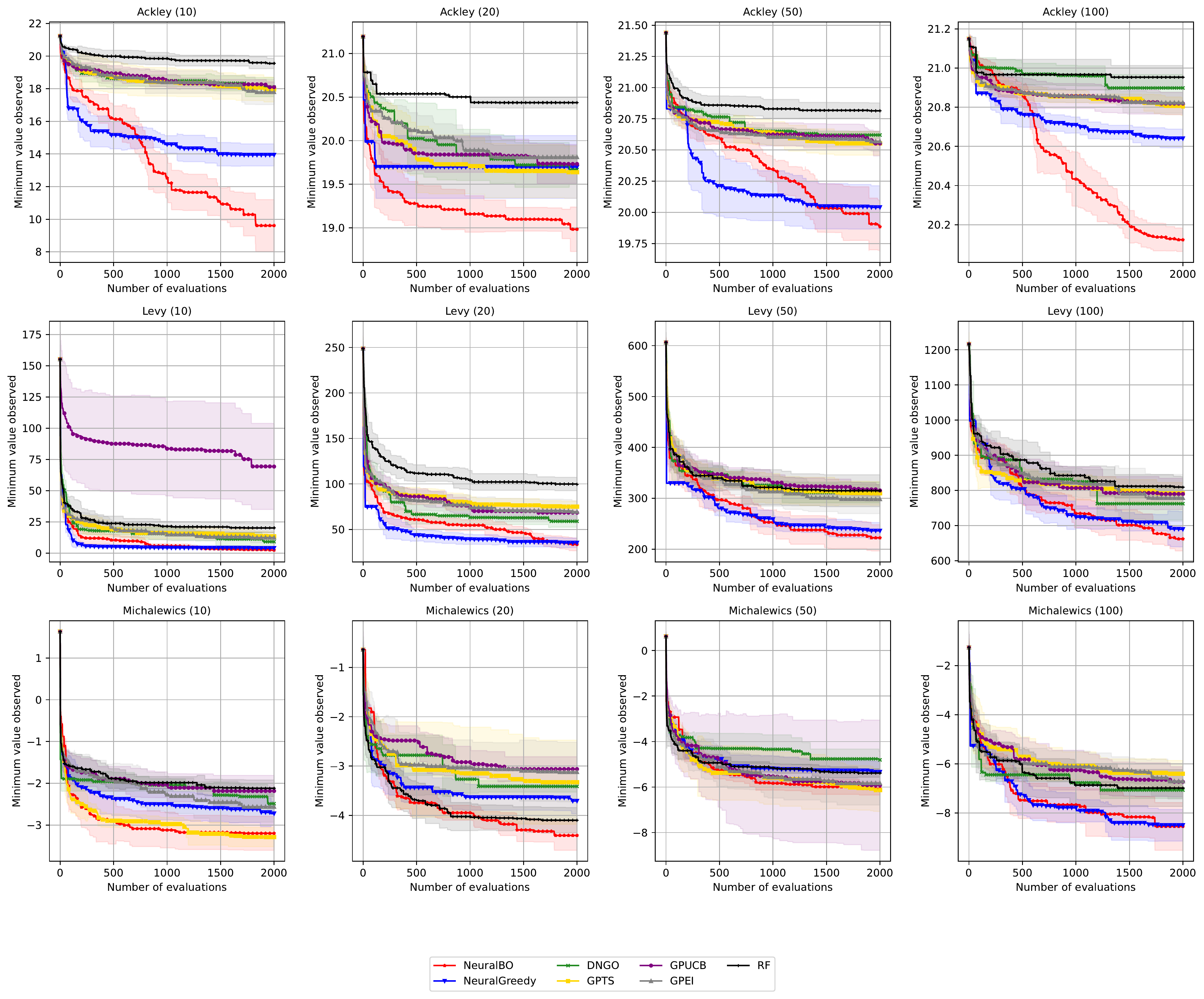}
    \caption{The plots show the minimum true value observed after optimizing several synthetic functions over 2000 iterations of our proposed algorithm and 6 baselines. The dimension of each function is shown in the parenthesis.}
    \label{fig:synthetic}
\end{figure}

All reported experiments are averaged over 10 runs, each using a random initialization. All the methods start with the same set of initial points. As seen from Figure \ref{fig:synthetic}, NeuralBO is comparable to or better than all other baselines, including GP-based BO algorithms, both NN-based BO algorithms (DNGO and NeuralGreedy) and one algorithm using random forest. Moreover, our algorithm is also promising for high dimensions.

In order to assess the statistical significance of whether our proposed method outperforms all baselines, we conducted a series of tests. First, a Kolmogorov-Smirnov (KS) test was performed to check if the sample sets for the hypothesis test follow normal distribution. The null hypothesis assumes no difference between the observed and normal distribution. The p-values obtained from the KS test are presented in Table \ref{tab:ks_test}.  As all p-values exceeded 0.05, we failed to reject the null hypothesis, indicating that all data can be considered as normally distributed. Subsequently, one-sided t-tests were employed (as the variance is unknown), along with the Benjamini-Hochberg correction for multiple hypotheses testing, for each pair of (baseline, NeuralBO). These tests aimed to determine whether the baseline achieves a lower objective function value than our proposed method, NeuralBO. The alternative hypothesis $H_a: \mu_\text{baseline} > \mu_{\text{Neural-BO}}$ was tested against the null hypothesis $H_0: \mu_\text{baseline} \le \mu_{\text{Neural-BO}}$. Detailed test results are provided in Table \ref{tab:t-test}, where each cell contains two values: the first value represents the p-value obtained from the t-test, and the second value (T or F) indicates the outcome of the Benjamini-Hochberg correction where "T" indicates that the null hypothesis is rejected, whereas an "F" indicates that it is not rejected.

\begin{table}[]
\resizebox{1 \textwidth}{!}{
\begin{tabular}{|cc|c|c|c|c|c|c|c|}
\hline
\multicolumn{2}{|c|}{}                                                                                          & \textbf{\begin{tabular}[c]{@{}c@{}}Neural\\ BO\end{tabular}} & \textbf{\begin{tabular}[c]{@{}c@{}}Neural\\ Greedy\end{tabular}} & \textbf{\begin{tabular}[c]{@{}c@{}}GP\\ UCB\end{tabular}} & \textbf{\begin{tabular}[c]{@{}c@{}}GP\\ TS\end{tabular}} & \textbf{\begin{tabular}[c]{@{}c@{}}GP\\ EI\end{tabular}} & \textbf{DNGO} & \textbf{RF} \\ \hline
\multicolumn{1}{|c|}{\multirow{4}{*}{\textbf{Ackley}}}                                                  & D=10  & 0.85                                                         & 0.83                                                             & 0.95                                                      & 0.95                                                     & 0.45                                                     & 0.23          & 0.79        \\ \cline{2-9} 
\multicolumn{1}{|c|}{}                                                                                  & D=20  & 0.31                                                         & 0.46                                                             & 0.91                                                      & 0.90                                                     & 1.00                                                     & 1.00          & 0.93        \\ \cline{2-9} 
\multicolumn{1}{|c|}{}                                                                                  & D=50  & 0.95                                                         & 0.38                                                             & 0.81                                                      & 0.99                                                     & 0.52                                                     & 0.85          & 0.67        \\ \cline{2-9} 
\multicolumn{1}{|c|}{}                                                                                  & D=100 & 0.84                                                         & 0.77                                                             & 0.94                                                      & 0.82                                                     & 0.82                                                     & 0.38          & 0.97        \\ \hline
\multicolumn{1}{|c|}{\multirow{4}{*}{\textbf{Levy}}}                                                    & D=10  & 0.75                                                         & 0.45                                                             & 0.79                                                      & 0.72                                                     & 0.72                                                     & 0.58          & 0.63        \\ \cline{2-9} 
\multicolumn{1}{|c|}{}                                                                                  & D=20  & 0.84                                                         & 1.00                                                             & 1.00                                                      & 0.79                                                     & 0.79                                                     & 0.91          & 0.89        \\ \cline{2-9} 
\multicolumn{1}{|c|}{}                                                                                  & D=50  & 0.76                                                         & 0.34                                                             & 0.75                                                      & 0.75                                                     & 0.98                                                     & 0.77          & 0.37        \\ \cline{2-9} 
\multicolumn{1}{|c|}{}                                                                                  & D=100 & 0.76                                                         & 0.59                                                             & 0.87                                                      & 0.90                                                     & 0.99                                                     & 0.93          & 0.95        \\ \hline
\multicolumn{1}{|c|}{\multirow{4}{*}{\textbf{\begin{tabular}[c]{@{}c@{}}Micha-\\ lewics\end{tabular}}}} & D=10  & 0.54                                                         & 0.56                                                             & 0.95                                                      & 0.88                                                     & 0.83                                                     & 0.77          & 0.55        \\ \cline{2-9} 
\multicolumn{1}{|c|}{}                                                                                  & D=20  & 0.60                                                         & 0.99                                                             & 0.43                                                      & 0.33                                                     & 0.31                                                     & 0.56          & 0.85        \\ \cline{2-9} 
\multicolumn{1}{|c|}{}                                                                                  & D=50  & 0.90                                                         & 0.80                                                             & 0.64                                                      & 0.39                                                     & 0.68                                                     & 0.99          & 0.95        \\ \cline{2-9} 
\multicolumn{1}{|c|}{}                                                                                  & D=100 & 0.70                                                         & 0.58                                                             & 0.96                                                      & 0.57                                                     & 0.96                                                     & 0.93          & 0.52        \\ \hline
\end{tabular}
}
\caption{The p-values of KS-test "whether the data obtained from running our methods Neural-BO and all baselines are normally distributed".}
\label{tab:ks_test}
\end{table}

\begin{table}[]
\resizebox{1 \textwidth}{!}{
\begin{tabular}{|cc|l|l|l|l|l|l|}
\hline
\multicolumn{2}{|c|}{}                                                                                          & \multicolumn{1}{c|}{\textbf{\begin{tabular}[c]{@{}c@{}}Neural\\ Greedy\end{tabular}}} & \multicolumn{1}{c|}{\textbf{GPUCB}} & \multicolumn{1}{c|}{\textbf{GPTS}} & \multicolumn{1}{c|}{\textbf{GPEI}} & \multicolumn{1}{c|}{\textbf{DNGO}} & \multicolumn{1}{c|}{\textbf{RF}} \\ \hline
\multicolumn{1}{|c|}{\multirow{4}{*}{\textbf{Ackley}}}                                                  & D=10  & (2.98e-07,   T)                                                                       & (3.11e-12, T)                       & (1.64e-11, T)                      & (2.3e-11, T)                       & (6.15e-09, T)                      & (2.09e-13, T)                    \\ \cline{2-8} 
\multicolumn{1}{|c|}{}                                                                                  & D=20  & (6.39e-05, T)                                                                         & (1.61e-06, T)                       & (4.72e-05, T)                      & (4.19e-08, T)                      & (0.000111, T)                      & (1.29e-05, T)                    \\ \cline{2-8} 
\multicolumn{1}{|c|}{}                                                                                  & D=50  & (0.0467, T)                                                                           & (2.23e-08, T)                       & (1.89e-08, T)                      & (8.5e-09, T)                       & (1.86e-08, T)                      & (2.11e-10, T)                    \\ \cline{2-8} 
\multicolumn{1}{|c|}{}                                                                                  & D=100 & (3.92e-14, T)                                                                         & (7.98e-16, T)                       & (2.43e-16, T)                      & (4.7e-16, T)                       & (6.44e-16, T)                      & (5.78e-17, T)                    \\ \hline
\multicolumn{1}{|c|}{\multirow{4}{*}{\textbf{Levy}}}                                                    & D=10  & (0.000171, T)                                                                         & (7.98e-06, T                        & (8.81e-09, T)                      & (6.6e-07, T)                       & (1.68e-05, T)                      & (1.62e-11, T)                    \\ \cline{2-8} 
\multicolumn{1}{|c|}{}                                                                                  & D=20  & (0.278, F)                                                                            & (3.28e-09, T)                       & (4.32e-11, T)                      & (1.8e-07, T)                       & (4.75e-05, T)                      & (8.01e-14, T)                    \\ \cline{2-8} 
\multicolumn{1}{|c|}{}                                                                                  & D=50  & (0.0971, F)                                                                           & (5.37e-09, T)                       & (2.72e-07, T)                      & (6.47e-06, T)                      & (0.000188, T)                      & (1.99e-06, T)                    \\ \cline{2-8} 
\multicolumn{1}{|c|}{}                                                                                  & D=100 & (0.096, F)                                                                            & (1.09e-06, T)                       & (4.85e-08, T)                      & (9.87e-05, T)                      & (0.00456, T)                       & (5.57e-07, T)                    \\ \hline
\multicolumn{1}{|c|}{\multirow{4}{*}{\textbf{\begin{tabular}[c]{@{}c@{}}Micha-\\ lewics\end{tabular}}}} & D=10  & (5.9e-06, T)                                                                          & (7.79e-08, T)                       & (0.472, F)                         & (1.68e-05, T)                      & (7.49e-06, T)                      & (1.87e-11, T)                    \\ \cline{2-8} 
\multicolumn{1}{|c|}{}                                                                                  & D=20  & (2.63e-05, T)                                                                         & (3e-09, T)                          & (0.00108, T)                       & (1.02e-05, T)                      & (0.000163, T)                      & (0.0161, T)                      \\ \cline{2-8} 
\multicolumn{1}{|c|}{}                                                                                  & D=50  & (0.000344, T)                                                                         & (3.47e-06, T)                       & (0.315, F)                         & (0.101, F)                         & (0.000871, T)                      & (0.000132, T)                    \\ \cline{2-8} 
\multicolumn{1}{|c|}{}                                                                                  & D=100 & (0.45, F)                                                                             & (8.2e-05, T)                        & (8.49e-06, T)                      & (0.000126, T)                      & (0.0404, T)                        & (0.00579, T)                     \\ \hline
\end{tabular}
}
\caption{One-sided t-tests were employed to assess whether the baseline achieves a lower function value compared to our proposed method, NeuralBO.  The null hypothesis $H_0: \mu_\text{baseline} \le \mu_{\text{Neural-BO}}$ and the alternative hypothesis:  $H_a: \mu_\text{baseline} > \mu_{\text{Neural-BO}}$. The p-value corresponding to each test is provided as the first value in each cell. Moreover, to account for multiple hypotheses testing, the Benjamini-Hochberg correction was applied and is reported as the second value in each cell. In the outcome, a "T" indicates that the null hypothesis is rejected, whereas an "F" signifies that it is not rejected.}
\label{tab:t-test}
\end{table}

Next, we used the COmparing Continuous Optimizers (COCO) framework\footnote{ \url{https://github.com/numbbo/coco}} to compare the performance of our method with baseline methods. COCO is a widely used benchmarking platform for continuous optimization algorithms, providing researchers with a standardized set of test functions to evaluate different optimization methods. The framework includes a collection of test functions that are designed to be challenging to optimize and exhibit various properties, such as multimodality, ill-conditioning, or weak global structure. To evaluate our algorithms, we used the BBOB test functions, which consist of 24 noiseless, single-objective, and scalable functions. The quality indicator needs to reach or exceed a predefined target value  $t$  for COCO to consider a single run over a problem successful. The performance of different optimizers is measured by the \textit{Fraction of function, target pairs}, which indicates the ratio of problems solved at optimization step $T$. For more details on how COCO measures the performance of different optimizers, please refer to \cite{hansen2021coco}. We set the number of evaluations for our optimizers to be 20 times the dimension of the problem. 

\begin{figure}[H]
\begin{subfigure}{.5\textwidth}
  \centering
  \includegraphics[width=\linewidth]{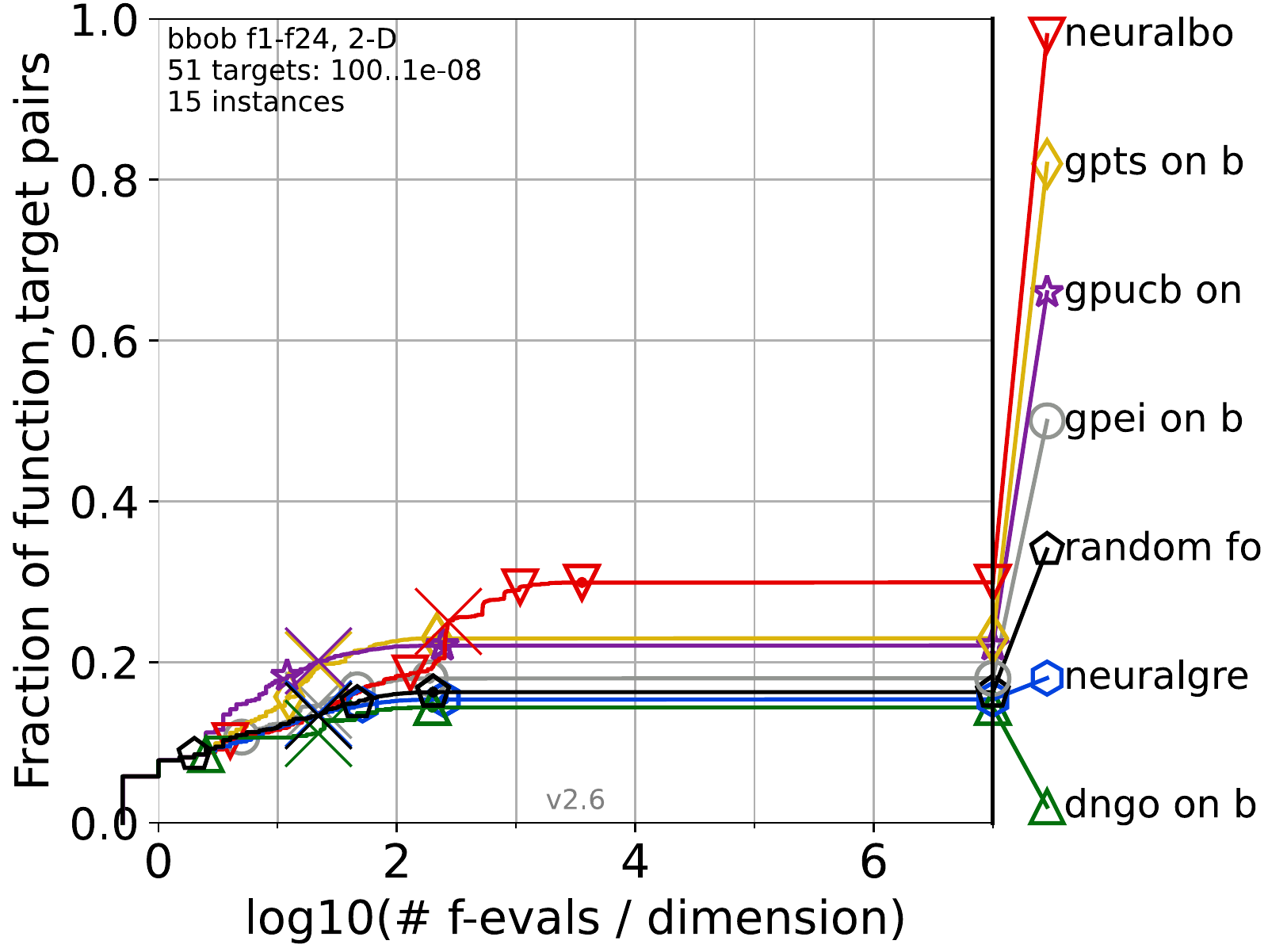}
  \caption{2D}
  \label{fig:coco_2d}
\end{subfigure}%
\begin{subfigure}{.5\textwidth}
  \centering
  \includegraphics[width=\linewidth]{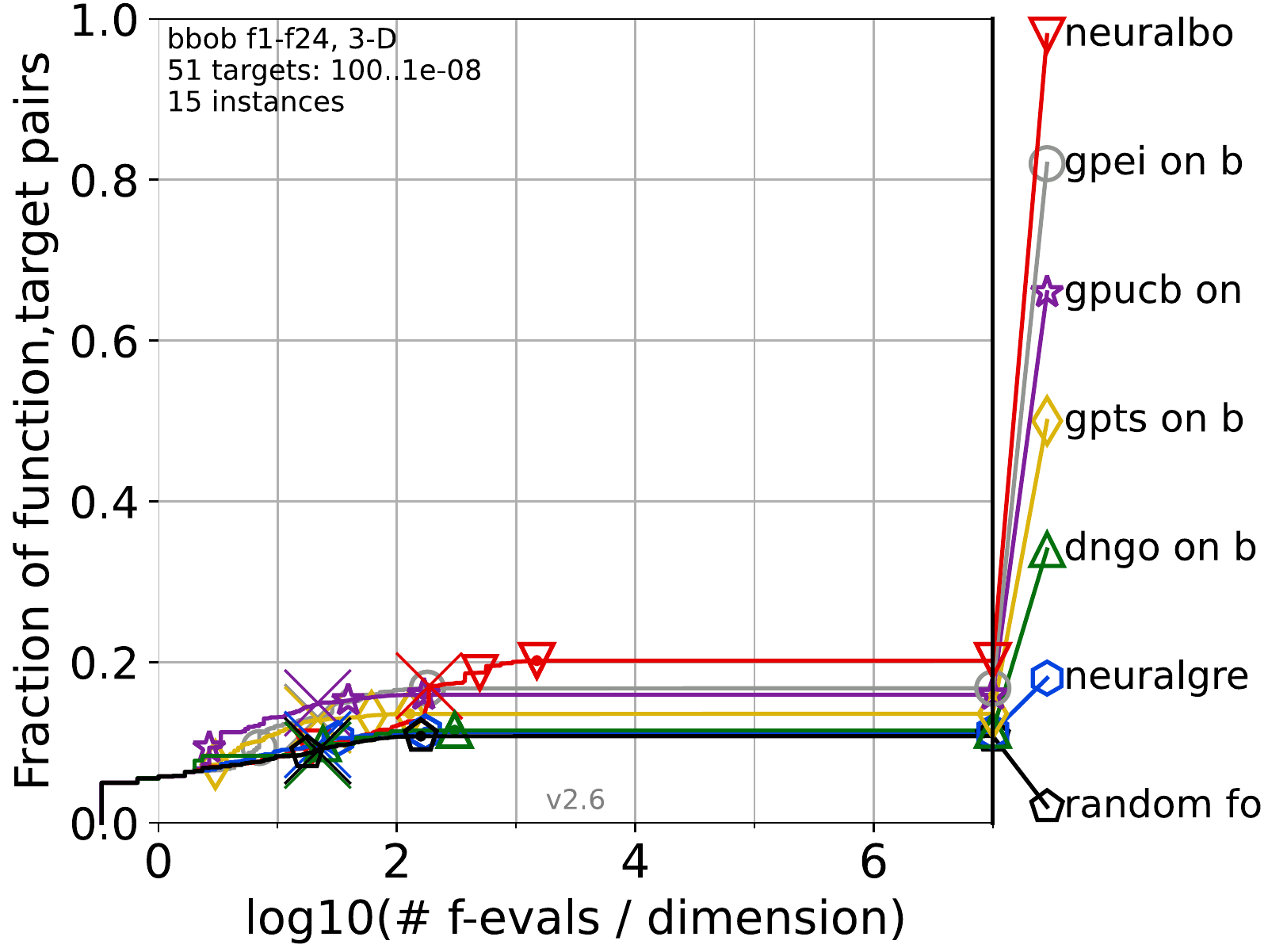}
  \caption{3D}
  \label{fig:coco_3d}
\end{subfigure}
\begin{subfigure}{.5\textwidth}
  \centering
  \includegraphics[width=\linewidth]{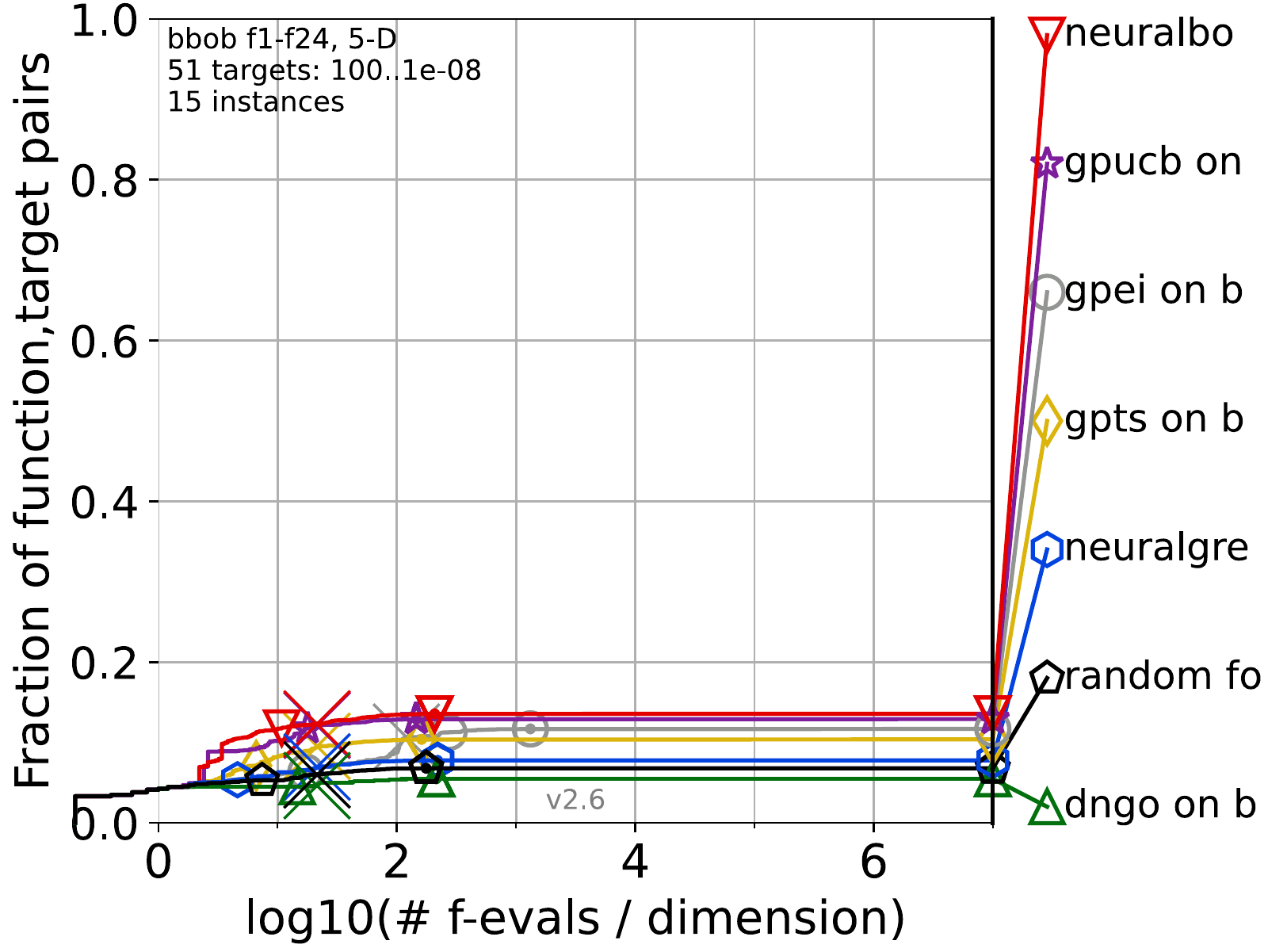}
  \caption{5D}
  \label{fig:coco_5d}
\end{subfigure}%
\begin{subfigure}{.5\textwidth}
  \centering
  \includegraphics[width=\linewidth]{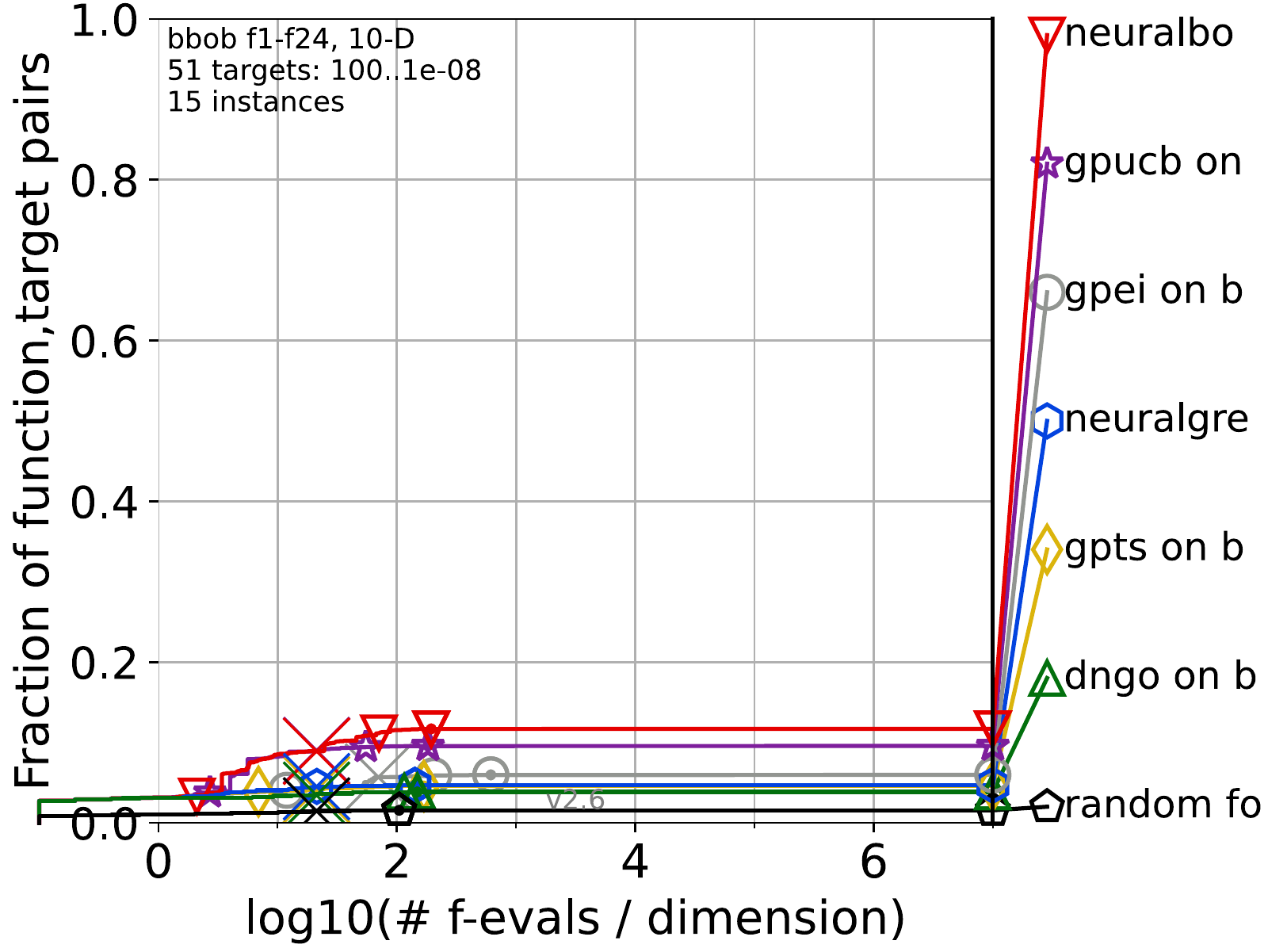}
  \caption{10D}
  \label{fig:coco_10d}
\end{subfigure}
\caption{The results of benchmarking our NeuralBO and the baselines with COCO framework on 24 BBOB noiseless objective functions with four different dimensions \{2,3,5,10\}.}
\label{fig:coco}
\end{figure}

In Figure \ref{fig:coco}, we present the findings of our experiment using the COCO benchmarking framework to evaluate all methods. The benchmarking comprised 24 noiseless functions with 15 instances, where each instance represented a unique condition of the function, such as the location of the optimum. Our method was found to outperform other baselines when assessed using the well-designed and publicly available COCO framework.

\subsection{Real Applications}
\subsubsection{Designing Sensitive Samples for Detection of Model Tampering}
\begin{figure}[H]
    \centering
    \includegraphics[width=\textwidth]{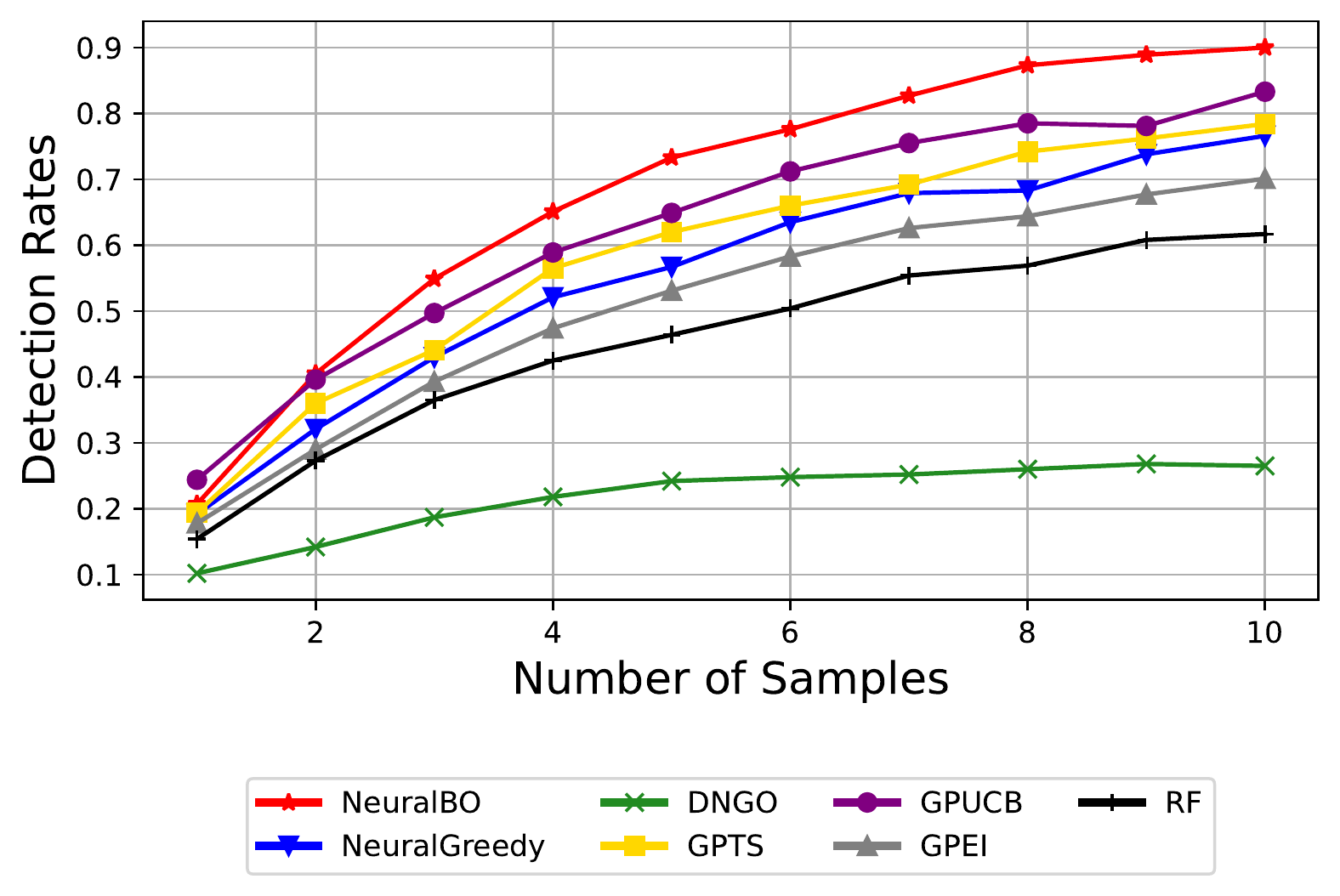}
    \caption{The plot shows the \textbf{detection rates} corresponding to the number of samples on MNIST dataset. The larger the number of sensitive samples, the higher the detection rate. As shown in the figure, NeuralBO can generate sensitive samples that achieve nearly 90\% of the detection rate with at least 8 samples.}
    \label{fig:sensitive_sample}
\end{figure}

We consider an attack scenario where a company offering Machine Learning as a service (MLaaS) hosts its model on a cloud. However, an adversary with backdoor access may tamper with the model to change its weight. This requires the detection of model tampering. To deal with this problem, \cite{he2018verideep} suggests generating a set of test vectors named \emph{Sensitive-Samples} $\{v_i\}_{i=1}^n$, whose outputs predicted by the compromised model will be different from the outputs predicted by the original model. As formalized in \cite{he2018verideep}, suppose we suspect a pre-trained model $f_\theta(x)$ of having been modified by an attacker after it was sent to the cloud service provider. Finding sensitive samples for verifying the model's integrity is equivalent to optimizing task: $v = \argmax_x \norm{\frac{\partial f_\theta(x)}{\partial \theta}}_F$, where $\norm{\cdot}_F$ is the Frobenius norm of a matrix. A \emph{successful detection} is defined as ``given $N_S$ sensitive samples, there is at least one sample, whose top-1 label predicted by the compromised model is different from the top-1 label predicted by the correct model''. Clearly, optimizing this expensive function requires a BO algorithm to be able to work with high-dimensional structured images, unlike usual inputs that take values in hyper-rectangles.

We used a hand-written digit classification model (pre-trained on MNIST data) as the original model and tampered it by randomly adding noise to each weight of this model. We repeat this procedure 1000 times to generate 1000 different tampered models. The top-1 accuracy of the MNIST original model is $93\%$ and is reduced to $87.73\% \pm 0.08\%$ after modifications.   

To reduce the computations, we reduce the dimension of images from $28 \times 28$ to $7 \times 7$ and do the optimization process in 49-dimensional space. After finding the optimal points, we recover these points to the original dimension by applying an upscale operator to get sensitive samples. We compare our method with all other baselines by the average detection rate with respect to the number of sensitive samples. From Figure \ref{fig:sensitive_sample}, it can be seen that our method can generate  samples with better detection ability than other baselines. This demonstrates the ability of  our method to deal with complex structured data such as images. 

\subsubsection{Unknown target document retrieval}
\begin{figure}[H]
    \centering
    \includegraphics[width=\textwidth]{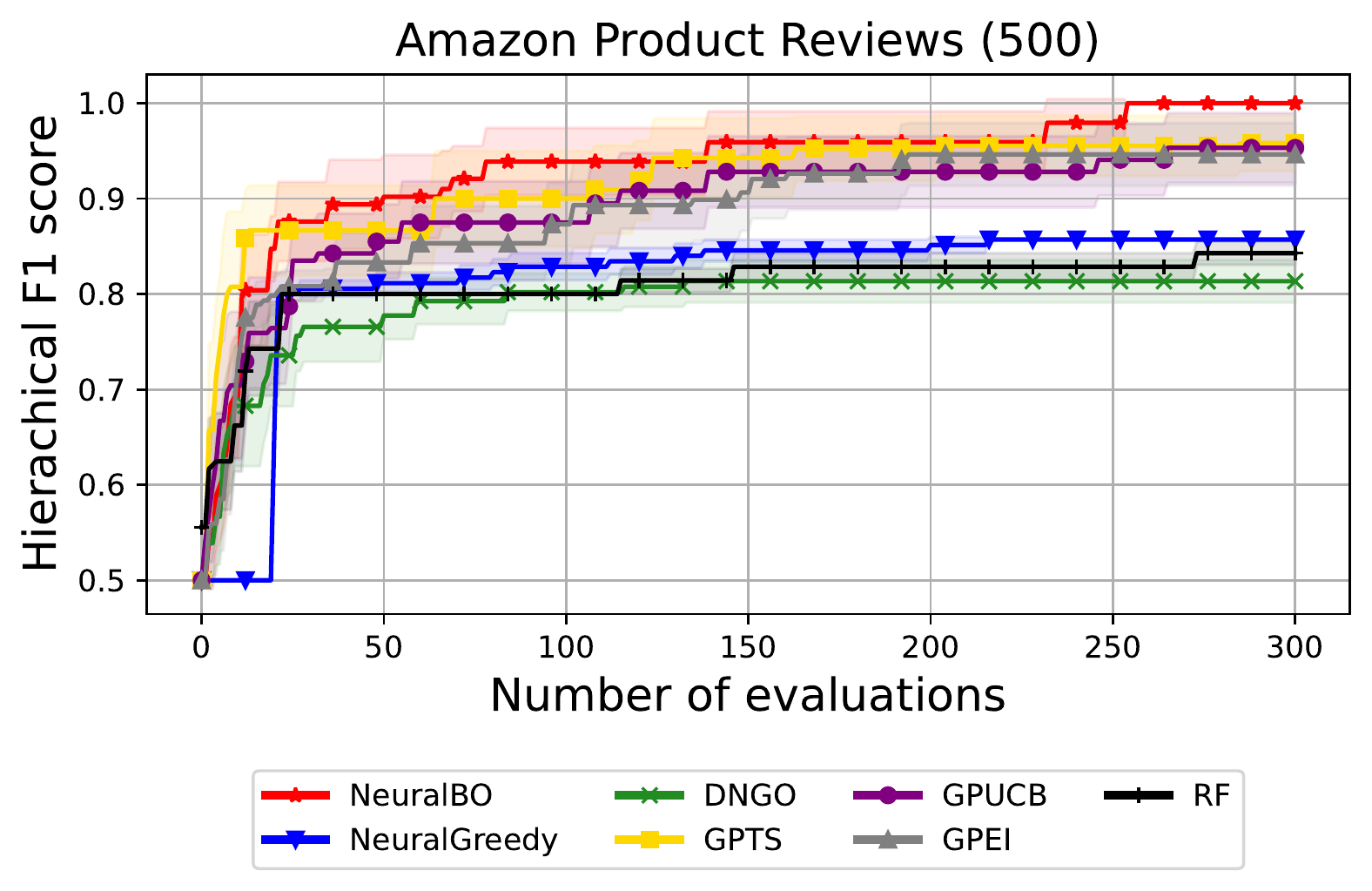}
    \caption{We search for the most related document for a specified target document in \textbf{Amazon product reviews} dataset and report the maximum \textbf{hierachical F1 score} found by all baselines. All methods show similar behaviour and NeuralBO performs comparably and much better than GP-based baselines.}
    \label{fig:text}
\end{figure}
Next, we evaluate our proposed method on a retrieval problem where our goal is to retrieve a document from a corpus that matches user's preference. The optimization algorithm works as follows: it retrieves a document, receives user's feedback score and then updates its belief and attempts again until it reaches a high score, or its budget is depleted. The objective target function is defined as the user’s evaluation of each document, which usually has a complex structure. It is clear that evaluations are expensive since the user must read each suggestion. Searching for the most relevant document is considered as finding document $d$ in the dataset $S_{text}$ that maximizes the match to the target document $d_t$: $d = \argmax_{d \in S_{text}} \textup{Match}(d, d_t)$, where $\textup{Match}(d, d_t)$ is a matching score between documents $d$ and $d_t$. We represent each document by a word frequency vector $x_{n} = (x_{n1}, \cdots, x_{nJ})$, where $x_{nj}$ is the number of occurrences of the $j$-th vocabulary term in the $n$-th document, and $J$ is the vocabulary size.  

Our experiment uses Amazon product reviews 
dataset \footnote{\url{https://www.kaggle.com/datasets/kashnitsky/hierarchical-text-classification}}, which are taken from users' product reviews from Amazon's online selling platform. This dataset has hierarchical categories, where the category classes are sorted from general to detail. The dataset are structured as: 6 classes in ``level 1'', 64 classes in ``level 2'' and 464 classes in ``level 3''. The number of users' reviews was originally 40K, which was reduced to 37738 after ignoring reviews with ``unknown" category. We choose the size of vocabulary to be 500 and use hierarchical F1-score introduced in  \cite{kiritchenko2005functional} as a scoring metric for the target and retrieved documents. We report the mean and standard deviation of hierarchical F1-score between target and retrieved documents over ten runs for all methods in Figure \ref{fig:text}. Figures \ref{fig:text} indicate that our method shows a better performance for the Amazon product review dataset in comparison with other approaches. 

\subsubsection{Optimizing control parameters for robot pushing} 
\begin{figure}[H]
    \centering
    \includegraphics[width=\textwidth]{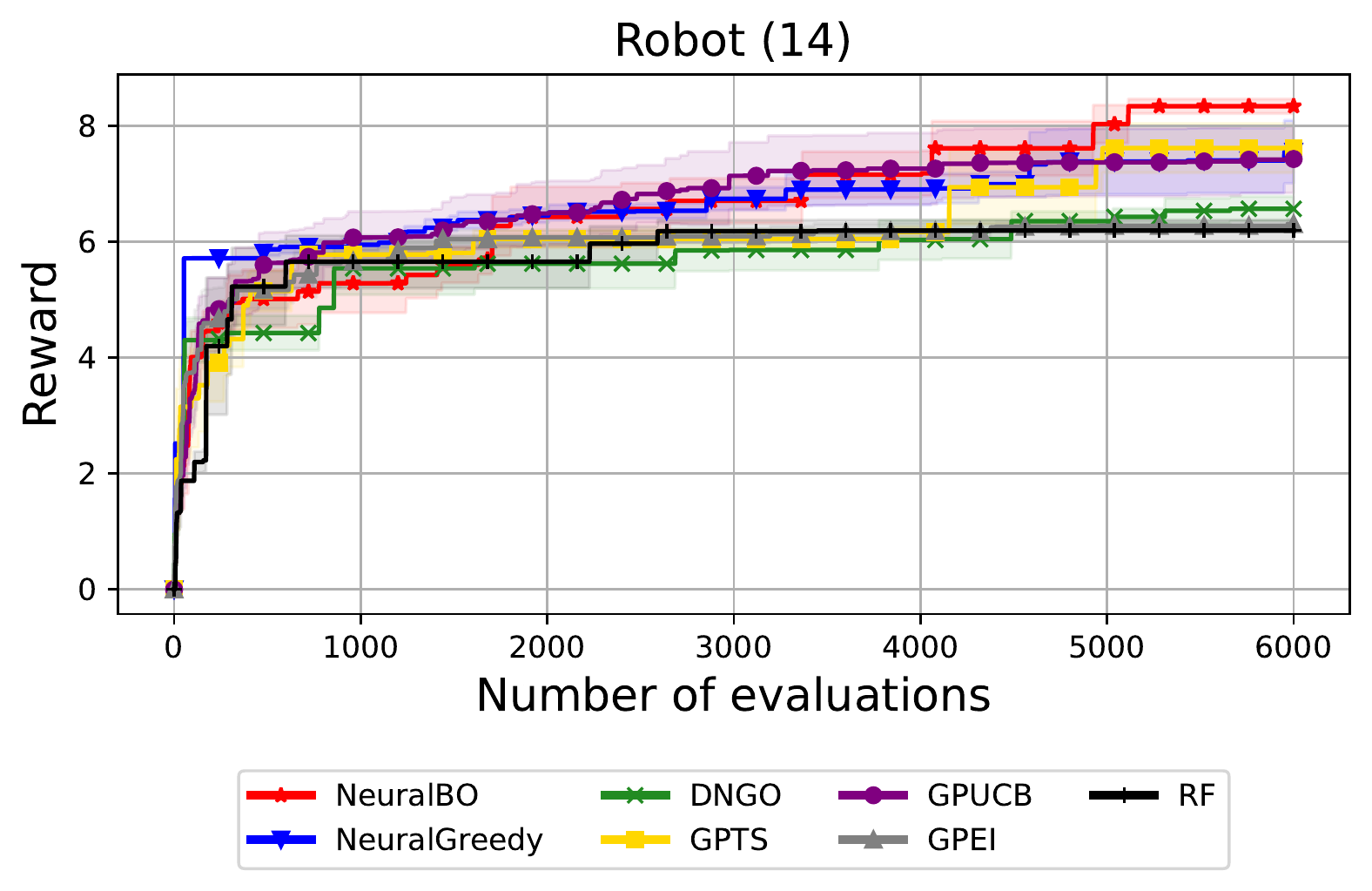}
    \vspace{-3mm}
    \caption{
    Optimization results for control parameters of 14D robot pushing problem. The X-axis shows iterations, and the y-axis shows the median of the best reward obtained.}
    \label{fig:robot_14D}
\end{figure}
\begin{figure}[t]
    \centering
    \includegraphics[width=\textwidth]{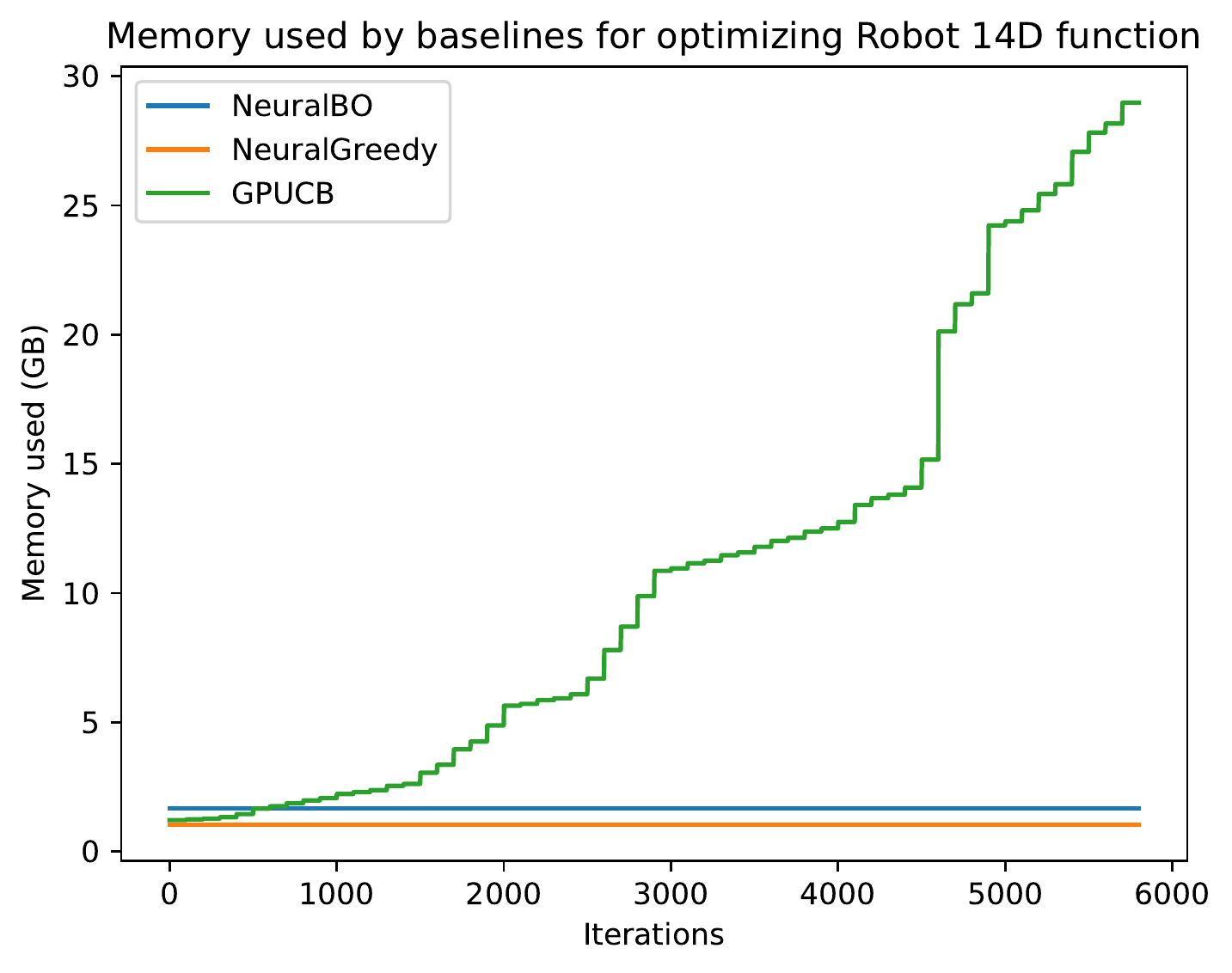}
    \vspace{-3mm}
    \caption{The GPU memory used by three practical baselines for 14D robot pushing control parameters optimization task.}
    \label{fig:test_mem_robot}
\end{figure}
Finally, we evaluate our method on tuning control parameters for robot pushing problem considered in \cite{wang2017max}. We run each method for a total of 6000 evaluations and repeat ten times to take average optimization results. NeuralBO and all other methods are initialized with 15 points. Figure \ref{fig:robot_14D} summarizes the median of the best rewards achieved by all methods. It can be seen that NeuralBO achieves the highest reward after 5K optimization steps. 

We also use this task to demonstrate the GPU memory used by three practical baselines: GP-UCB, NeuralBO, and NeuralGreedy in Figure  \ref{fig:test_mem_robot}. The NeuralGreedy and NeuralBO GPU memory requirements do not appear to have changed as the memory used by these two methods remains almost constant, unlike GP. For GP, GPU memory requirement grows fast with the iterations due to the need to compute the inverse of larger and larger covariance matrices. We report here GP-UCB, but similar behavior is also noted for GP-EI and GP-TS. Neural Greedy is slightly lesser memory-consuming because it only needs memory to store the neural network model and for the training process. NeuralBO used a little bit more memory due to the calculation of matrix $\mathbf{U}$ described in Algorithm \ref{alg:NeuralBO}. However, this matrix has a fixed size; therefore, the memory consumption does not increase over time. We do not report memory used by RF and DNGO because their implementations are CPU-based, and their running speeds are too slow compared to all the above baselines.

\section{Conclusion}
We proposed a new algorithm for Bayesian optimization using deep neural networks. A key advantage of our algorithm is that it is computationally efficient and performs better than traditional Gaussian process based methods, especially for complex structured design problems. We provided rigorous theoretical analysis for our proposed algorithm and showed that its cumulative regret converges with a sub-linear regret bound. Using both synthetic benchmark optimization functions and a few real-world optimization tasks, we showed the effectiveness of our proposed algorithm. 

Our proposed method currently employs over-parametric neural networks, which can be analyzed using neural tangent kernels. As a future work, it would be interesting to investigate if it is possible to extend this work beyond over-parametric neural networks to allow a broader class of neural networks that may have smaller widths.   
\section{Acknowledgement}
This research was partially supported by the Australian Government through the Australian Research Council's Discovery Projects funding scheme (project DP210102798). The views expressed herein are those of the authors and are not necessarily those of the Australian Government or Australian Research Council.
\bibliographystyle{elsarticle-num} 
\bibliography{references.bib}
\newpage
\appendix

\centerline{ \textbf{\huge Appendix}}
\vspace{3cm}
In this part, we provide the proof for Theorem \ref{theorem:main}. The main lemmas are from Lemma \ref{lemma:bound_hil} to Lemma \ref{lemma:min_sigma}. Some main lemmas require auxiliary lemmas to complete their proofs. These auxiliary lemmas are provided instantly after the main lemmas and later proved in \ref{sec:aux_appendix}. 

\section{}
\label{sec:main_appendix}
To begin, we consider the following condition of the neural network width $m$
\begin{condition}
\label{cond:set}
The network width $m$ satisfies
  \begin{align*}
     m & \geq C \max \Big\{ \sqrt{\lambda} L^{-3/2} [\log (TL^2 \alpha)]^{3/2}, T^6 L^6 \log (TL/ \alpha) \max \{\lambda_0^{-4},1 \} ] \Big\} 
     \\
     m  & [\log m ]^{-3} \geq CTL^{12} \lambda^{-1} + CT^7 \lambda^{-8}L^{18}(\lambda + LT)^6 + CL^{21}T^7\lambda^{-7}(1 + \sqrt{T/ \lambda})^6,
    \end{align*} 
    where $C$ is a positive absolute constant.
\end{condition}
Under this flexible condition of the search space as mentioned in \ref{sec:regret_analysis}, we need some new results. Following \cite{allen2019convergence}, we first define \[\mathbf{h}_{i,0} = \mathbf{x}, \mathbf{h}_{i,l} = \phi(\mathbf{W}_l \mathbf{h}_{i, l-1}), l \in [L]\]
as the output of the $l$-th hidden layer. With this definition, we provide a norm bound of $\mathbf{h}_{i, l-1}$ as follows.
\begin{lemma}[Lemma 7.1, \cite{allen2019convergence}]
\label{lemma:bound_hil}
If $\epsilon \in [0,1]$,  with probability at least $1-\mathcal{O}(nL)e^{-\Omega(m\epsilon^2/L)}$, we have
\[ \forall i \in [T], l \in [L], \norm{\mathbf{h}_{i, l-1}} \in [ae^{-\epsilon}, be^\epsilon] 
\]
\end{lemma}


The following lemma is the concentration property of sampling value $\widetilde{f}_t(\mathbf{x})$ from estimated mean value $h(\mathbf{x}; \boldsymbol{\theta}_{t-1})$.
\begin{lemma}
\label{lemma:sampling_bound}
For any $t \in [T]$, and any finite subset $\mathcal{D}_t \subset\ \mathcal{D}$,  pick $c_t =  \sqrt{4\log t + 2 \log \ \lvert \mathcal D_t \rvert}$. Then  
\[\lvert \widetilde{f}_t(\mathbf{x}) - h(\mathbf{x}; \boldsymbol{\theta}_{t-1}) \rvert \leq c_t\nu_t \sigma_t(\mathbf{x}), \forall \mathbf{x} \in \mathcal{D}_t, \] 
holds with probability $\geq  1-t^{-2}$.
\end{lemma}
To prove Lemma \ref{lemma:sampling_bound}, we need a concentration bound on Gaussian distributions \cite{hoffman2013exploiting} as follows: 
\begin{sublemma}[\cite{hoffman2013exploiting}]
\label{lemma:gauss_concetration}
Consider a normally distributed random variable $X \sim \mathcal{N}(\mu, \sigma^2)$ and $\beta \geq 0$. The probability that $X$ is within a radius of $\beta \sigma$ from its mean can then be written as:
\[ \mathbb{P}(\lvert X - \mu\rvert \leq \beta  \sigma\ ) \geq 1-\exp(\beta^2/2)\]
\end{sublemma}
\newproof{pl1}{Proof of Lemma \ref{lemma:sampling_bound}}
\begin{pl1}
Because the sampled value $\widetilde{f}_t(\mathbf{x})$ is sampled from  $\mathcal{N}(h(\mathbf{x}; \boldsymbol{\theta}_{t-1}), \nu_t^2\sigma_t^2(\mathbf{x}))$, applying the concentration property in Lemma \ref{lemma:gauss_concetration}, we have:

\[\mathbb{P}(\rvert \widetilde{f}_t(\mathbf{x}) -h(\mathbf{x}; \boldsymbol{\theta}_{t-1}) \lvert \leq c_t \nu_t \sigma_t(\mathbf{x}) \lvert \mathcal{F}_{t-1}) \geq 1-\exp(-c_t^2/2)
\]
Taking union bound over $\mathcal{D}_t$, we have for any $t$:
\[\mathbb{P}(\rvert \widetilde{f}_t(\mathbf{x}) - h(\mathbf{x}; \boldsymbol{\theta}_{t-1}) \lvert \leq c_t \nu_t \sigma_t(\mathbf{x}) \lvert \mathcal{F}_{t-1}) \geq 1-\rvert \mathcal{D}_t \rvert\exp(-c_t^2/2) \]
Picking $c_t = \sqrt{4 \log t + 2 \log \lvert \mathcal{D}_t \rvert}$, we get the bound:

\[ \mathbb{P}(\rvert \widetilde{f}_t(\mathbf{x}) - h(\mathbf{x}; \boldsymbol{\theta}_{t-1}) \lvert \leq c_t \nu_t \sigma_t(\mathbf{x}) \lvert \mathcal{F}_{t-1}) \geq 1-\frac{1}{t^2}\]
\end{pl1}

By combining Lemma \ref{lemma:bound_hil} with techniques used in Lemma 4.1 of \cite{cao2019generalization}, Lemma B.3 
 of \cite{cao2019generalization} and Theorem 5 
of \cite{allen2019convergence}, we achieve a concentration property of estimated value $h(\mathbf{x}; \boldsymbol{\theta}_{t-1})$ from its true value $f(\mathbf{x})$ as follows:
\begin{lemma}
\label{lemma:predictive_bound}
Suppose the width of the neural network $m$ satisfies Condition \ref{cond:set}. Given any $\alpha \in (0,1)$ and set $\eta = C(m\lambda + mLT)^{-1}$.  Then for any $t \in [T]$, we have

\[\lvert h(\mathbf{x}; \boldsymbol{\theta}_{t-1}) - f(\mathbf{x}) \rvert \leq \nu_t \sigma_t(\mathbf{x}) + \epsilon(m), \forall \mathbf{x} \in \mathcal{D}_t, \] holds with probability $\geq  1-\alpha/2$ and 
\begin{equation*}
    \begin{split}
     \epsilon(m) = & \frac{b}{a} C_{\epsilon,1} m^{-1/6}\lambda^{-2/3}L^3 \sqrt{\log m} + \frac{b}{a} C_{\epsilon,2} (1-\eta m\lambda)^J \sqrt{TL/\lambda}\\
     & + \left(\frac{b}{a}\right)^3 C_{\epsilon,3} m^{-1/6} \sqrt{\log m} L^4 T^{5/3} \lambda^{-5/3} (1+\sqrt{T/\lambda}), \\
    \end{split}
\end{equation*}
where $\{C_{\epsilon,i}\}_{i=1}^3$ are positive constants and $a,b$ is lower and upper norm bound of input $\mathbf{x}$ as assumed in Section \ref{sec:problem_setting}.
\end{lemma}
First, we define some necessary notations about linear and kernelized models.  
\begin{appendixdefinition}
\label{def:linear_kernelized_terms}
Let us define terms for the convenience as follows:
\begin{equation*}
\begin{split}
\mathbf{G}_t & = (\mathbf{g}(\mathbf{x}_1; \boldsymbol{\theta}_0),\cdots,\mathbf{g}(\mathbf{x}_t; \boldsymbol{\theta}_0))\\
     \mathbf{f}_t & = (f(\mathbf{x}_1), \cdots, f(\mathbf{x}_t))^\top \\
     \mathbf{y}_t & = (y_1, \cdots, y_t)^\top \\
     \boldsymbol{\epsilon}_t & = (f(\mathbf{x}_1) - y_1, \cdots, f(\mathbf{x}_t) - y_t)^\top,
\end{split}
\end{equation*}
\end{appendixdefinition}
where $\boldsymbol{\epsilon}_t$ is the reward noise at time $t$. We recall that the definition of $\boldsymbol{\epsilon}_t$ is simply from our setting in Section \ref{sec:problem_setting}. It can be verified that $\mathbf{U}_t = \lambda\mathbf{I} + \mathbf{G}_t\mathbf{G}_t^\top/m$. 
We also define $\mathbf{K}_t = \lambda\mathbf{I} + \mathbf{G}_t ^\top\mathbf{G}_t /m$. 
We next reuse a lemma from \cite{zhang2021neural} to bound the difference between the outputs of the neural network and the linearized model.

\begin{sublemma} 
\label{lemma:NN_vs_linear}
Suppose the network width $m$ satisfies Condition \ref{cond:set}.
Then, set $\eta = C_1(m\lambda + mLT)^{-1}$, with probability at least $1 - \alpha$ over the random initialization of $\boldsymbol{\theta}_0$, we have $\forall \mathbf{x} \in \mathcal{D}, 0 < a \leq \norm{\bf x}_2  \leq b$
\begin{equation*}
\begin{split}
    & \left \lvert h(\mathbf{x}; \boldsymbol{\theta}_{t-1})- \langle \mathbf{g}(\mathbf{x}; \boldsymbol{\theta}_0); \mathbf{G}_{t-1}^\top \mathbf{U}^{-1}_{t-1} \mathbf{r}_{t-1}/m \rangle \right\rvert \\ 
    \leq &  \frac{b}{a} C_{\epsilon,1} t^{2/3}m^{-1/6} \lambda^{-2/3} L^3 \sqrt{\log m} + \frac{b}{a} C_{\epsilon,2}(1 - \eta m \lambda)^J \sqrt{tL/ \lambda}\\
    & + \left(\frac{b}{a} \right)^3 C_{\epsilon,3} m^{-1/6} \sqrt{\log m} L^4 t^{5/3} \lambda ^{-5/3} \left(1 + \sqrt{t/\lambda} \right),
\end{split}
\end{equation*}
where $\{C_{\epsilon,i}\}_{i=1}^3$ are positive constants. We provide the proof for this lemma in \ref{NN_vs_linear_proof}.
\end{sublemma}

\begin{sublemma}
\label{lemma:noise_affeted_bound}
Let $\alpha \in (0,1)$. Recall that matrix $\mathbf{U}_{t-1}$ is defined in Algorithm \ref{alg:NeuralBO}, $\mathbf{G}_{t-1}$ is defined in Definition \ref{def:linear_kernelized_terms} and $\mathcal{\epsilon}_{t-1}$ is i.i.d sub-Gaussian  observation noises up to optimization iteration $t-1$, with parameter $R$. Then with probability  $1-\alpha$, we have
\[ \left\lvert \mathbf{g}(\mathbf{x}; \boldsymbol{\theta}_0)^\top \mathbf{U}^{-1}_{t-1} \mathbf{G}_{t-1} \boldsymbol{\epsilon}_{t-1}/m   \right\rvert \leq \sigma_t(\mathbf{x})\frac{R}{\sqrt{\lambda}} \sqrt{2 \log(\frac{1}{\alpha})}\]
\end{sublemma}
The proof for Lemma \ref{lemma:noise_affeted_bound} is given in \ref{noise_affeted_bound_proof}. Then, the following lemma provides the upper bound for approximating the NTK with the empirical gram matrix of the neural network at initialization by the maximum information gain associated with the NTK. 

\begin{sublemma}
\label{lemma:log_det_Kt_bound}
Let $\alpha \in (0,1)$. If the network width $m$ satisfies  $m \geq C T^6L^6 \log(TL/\alpha)$, then with probability at least $1-\alpha$, with the points chosen from Algorithm \ref{alg:NeuralBO}, the following holds for every $t \in [T]$:
\[ \log \det (\mathbf{I} + \lambda^{-1} \mathbf{K}_t) \le 2\gamma_t + 1,\]
where $\gamma_t$ is maximum information gain associated with the kernel $k_\textup{NTK}$. We provided the proof of Lemma \ref{lemma:log_det_Kt_bound} in \ref{log_det_Kt_bound_proof}.
\end{sublemma}

\begin{sublemma}[Lemma D.2, \cite{kassraie2022neural}] 
\label{lemma:RKHS_expression}
Let $\alpha \in (0,1)$. Under Assumption \ref{assumption:sufficient_exploration} , if the network width $m$ satisfies  $m \geq C T^6L^6 \log(TL/\alpha)$ and $f$ be a member of $\mathcal{H}_{k_\text{NTK}}$ with bounded RKHS norm $\norm{f}_{k_\text{NTK}} \leq B$, then with probability at least $1-\alpha$, there exists $\mathbf{w} \in \mathbb{R}^p$ such that 
\[ f(\mathbf{x}) = \langle \mathbf{g}(\mathbf{x}; \boldsymbol{\theta}_0), \mathbf{w} \rangle, \norm{\mathbf{w}}_2 \leq \sqrt{\frac{2}{m}}B \]
\end{sublemma}
We remark that in our proofs, we assume $k_{\text{NTK}}(\mathbf{x}, \mathbf{x}) \leq 1$ for simplicity. Now we are going on to prove Lemma \ref{lemma:predictive_bound}
\newproof{pl2}{Proof of Lemma \ref{lemma:predictive_bound}}
\begin{pl2}
First of all, since m satisfies Condition \ref{cond:set}, then with the choice of $\eta$, the condition required in Lemma \ref{lemma:NN_vs_linear} - \ref{lemma:log_det_Kt_bound} are satisfied. Thus, taking a union bound, we have with probability at least $1 - 3\alpha$, that the bounds provided by these lemmas hold. 
As we assume that $f$ is in RKHS $\mathcal{H}_{k_\textup{NTK}}$ with NTK kernel, and $\mathbf{g}(\mathbf{x}; \boldsymbol{\theta}_0)/\sqrt{m}$  can be considered as finite approximation of $\varphi(\cdot)$, the feature map of the NTK from $\mathbb{R}^d \rightarrow \mathcal{H}_{k_\textup{NTK}}$. From Lemma \ref{lemma:RKHS_expression} , there exists $\mathbf{w} \in \mathbb{R}^p$ such that $f(\mathbf{x}) = \langle \mathbf{g}(\mathbf{x}; \boldsymbol{\theta}_0), \mathbf{w} \rangle = \mathbf{g}(\mathbf{x}; \boldsymbol{\theta}_0)^\top \mathbf{w}$. 
Then for any $t \in [T]$, we will first provide the difference between the target function and the linear function
$\langle \mathbf{g}(\mathbf{x}; \boldsymbol{\theta}_0); \mathbf{U}^{-1}_{t-1} \mathbf{G}_{t-1} \mathbf{r}_{t-1}/m \rangle$ as:
\begin{equation}
\label{ieqn:confidence_interval}
    \begin{split}
         & \left \lvert f(\mathbf{x}) - \langle \mathbf{g}(\mathbf{x}; \boldsymbol{\theta}_0); \mathbf{U}^{-1}_{t-1} \mathbf{G}_{t-1} \mathbf{r}_{t-1}/m \rangle   \right \rvert  \\
        & = \left\lvert f(\mathbf{x}) - \mathbf{g}(\mathbf{x}; \boldsymbol{\theta}_0)^\top  \mathbf{U}^{-1}_{t-1} \mathbf{G}_{t-1} \mathbf{r}_{t-1}/m \right\rvert \\
        & \leq \left\lvert f(\mathbf{x}) - \mathbf{g}(\mathbf{x}; \boldsymbol{\theta}_0)^\top  \mathbf{U}^{-1}_{t-1}
        \mathbf{G}_{t-1}\mathbf{f}_{t-1}/m \right\rvert + 
        \left\lvert \mathbf{g}(\mathbf{x}; \boldsymbol{\theta}_0)^\top \mathbf{U}^{-1}_{t-1}
        \mathbf{G}_{t-1} \boldsymbol{\epsilon}_{t-1}/m \right\rvert\\
        & = \left\lvert \mathbf{g}(\mathbf{x}; \boldsymbol{\theta}_0)^\top \mathbf{w} - \mathbf{g}(\mathbf{x}; \boldsymbol{\theta}_0)^\top  \mathbf{U}^{-1}_{t-1} 
        \mathbf{G}_{t-1}
        \mathbf{G}_{t-1}^\top \mathbf{w}/m \rangle \right\rvert + 
        \left\rvert  \mathbf{g}(\mathbf{x}; \boldsymbol{\theta}_0)^\top \mathbf{U}^{-1}_{t-1} \mathbf{G}_{t-1} \boldsymbol{\epsilon}_{t-1}/m  \right\rvert\\
        & = \left\lvert \mathbf{g}(\mathbf{x}; \boldsymbol{\theta}_0)^\top \left( \mathbf{I} -  \mathbf{U}^{-1}_{t-1}  \mathbf{G}_{t-1} \mathbf{G}_{t-1}^\top/m  \right) \mathbf{w}  \right \vert + 
        \left\lvert  \mathbf{g}(\mathbf{x}; \boldsymbol{\theta}_0)^\top \mathbf{U}^{-1}_{t-1} \mathbf{G}_{t-1} \boldsymbol{\epsilon}_{t-1}/m  \right\rvert \\
        & = \left\lvert \mathbf{g}(\mathbf{x}; \boldsymbol{\theta}_0)^\top \left( \mathbf{I} -  \mathbf{U}^{-1}_{t-1} \left( \mathbf{U}_{t-1} -\lambda \mathbf{I} \right)  \right) \mathbf{w}  \right \vert +
        \left \lvert  \mathbf{g}(\mathbf{x}; \boldsymbol{\theta}_0)^\top \mathbf{U}^{-1}_{t-1} \mathbf{G}_{t-1} \boldsymbol{\epsilon}_{t-1}/m  \right \rvert \\
        & = \left\lvert \lambda \mathbf{g}(\mathbf{x}; \boldsymbol{\theta}_0)^\top \mathbf{U}^{-1}_{t-1} \mathbf{w}  \right\rvert  + \left\lvert \mathbf{g}(\mathbf{x}; \boldsymbol{\theta}_0)^\top \mathbf{U}^{-1}_{t-1} \mathbf{G}_{t-1} \boldsymbol{\epsilon}_{t-1}/m   \right\rvert \\
        & \leq \norm{\mathbf{w}}_{k_{\text{NTK}}}  \norm{ \lambda  \mathbf{U}^{-1}_{t-1} \mathbf{g}(\mathbf{x}; \boldsymbol{\theta}_0)}_{k_{\text{NTK}}} + \left\lvert \mathbf{g}(\mathbf{x}; \boldsymbol{\theta}_0)^\top \mathbf{U}^{-1}_{t-1} \mathbf{G}_{t-1} \boldsymbol{\epsilon}_{t-1}/m   \right\rvert \\
        & \leq  \norm{\mathbf{w}}_{k_{\text{NTK}}}  \sqrt {\lambda \mathbf{g}(\mathbf{x}; \boldsymbol{\theta}_0)^\top \mathbf{U}^{-1}_{t-1} \mathbf{g}(\mathbf{x}; \boldsymbol{\theta}_0)}  + \left\lvert \mathbf{g}(\mathbf{x}; \boldsymbol{\theta}_0)^\top \mathbf{U}^{-1}_{t-1} \mathbf{G}_{t-1} \boldsymbol{\epsilon}_{t-1}/m   \right\rvert \\
        & \leq \sqrt{2}B \sigma_t(\mathbf{x}) + \sigma_t(\mathbf{x})\frac{R}{\sqrt{\lambda}} \sqrt{2 \log(\frac{1}{\alpha})} \\
    \end{split}
\end{equation}
where the first inequality uses triangle inequality and the fact that $\mathbf{r}_{t-1}= \mathbf{f}_{t-1} + \boldsymbol{\epsilon}_{t-1}$. The second inequality is from the reproducing property of function relying on RKHS, and the fourth equality is from the verification noted in Definition  \ref{def:linear_kernelized_terms}. The last inequality directly uses the results from Lemma \ref{lemma:noise_affeted_bound} and Lemma \ref{lemma:RKHS_expression}. We have a more compact form of the inequality \ref{ieqn:confidence_interval} as: 
\[\rvert f(\mathbf{x}) - \langle \mathbf{g}(\mathbf{x}; \boldsymbol{\theta}_0); \mathbf{U}^{-1}_{t-1} \mathbf{G}_{t-1} \mathbf{r}_{t-1}/m \rangle \lvert \leq \nu_t \sigma_t (\mathbf{x}), 
\]
where we set $\nu_t = \sqrt{2}B + \frac{R}{\sqrt{\lambda}}\sqrt{2 \log(1/ \alpha)}$. 
Then, by combining this bound with
Lemma \ref{lemma:NN_vs_linear}, we have

\begin{equation*}
    \begin{split}
        \lvert h(\mathbf{x}; \boldsymbol{\theta}_{t-1}) - f(\mathbf{x}) \rvert & \leq \nu_t \sigma_t (\mathbf{x}) + \frac{b}{a} C_{\epsilon,1} t^{2/3}m^{-1/6} \lambda^{-2/3} L^3 \\ &
        \quad + \frac{b}{a} C_{\epsilon,2}(1 - \eta m \lambda)^J \sqrt{tL/\lambda} \\ &
        \quad +  \left(\frac{b}{a} \right)^3 C_{\epsilon,3} m^{-1/6} \sqrt{\log m} L^4 t^{5/3} \lambda ^{-5/3} \left(1 + \sqrt{t/\lambda}  \right) \\
        & \leq \nu_t \sigma_t (\mathbf{x}) + \epsilon(m)
    \end{split}
\end{equation*}
By setting $\alpha $ to $\alpha/3$ (required by the union bound discussed at the beginning of the proof) and taking $t=T$, we get the result presented in Lemma \ref{lemma:predictive_bound}.
\end{pl2}

The next lemma gives a lower bound of the probability that the sampled value $\widetilde{f}_t (\mathbf{x})$ is larger than the true function value up to the approximation error $\epsilon(m)$.
\begin{lemma}
\label{lemma:sampled_value_vs_real_value}
For any $t \in [T], \mathbf{x} \in D$, we have $\mathbb{P}(\widetilde{f}_t (\mathbf{x}) + \epsilon(m) > f(\mathbf{x})) \geq (4e\pi)^{-1}$.
\end{lemma}

\newproof{pl3}{Proof of Lemma \ref{lemma:sampled_value_vs_real_value}}
\begin{pl3}
Following proof style of Lemma 8 \cite{zhou2020neural}, using Lemma \ref{lemma:predictive_bound} and Gaussian anti-concentration property, we have 
\begin{equation*}
    \begin{split}
        & \mathbb{P}(\widetilde{f_t}(\mathbf{x}) + \epsilon(m) > f(\mathbf{x}) \lvert \mathcal{F}_{t-1}) \\
        = & \mathbb{P} \Bigg(\frac{\widetilde{f_t}(\mathbf{x}) - h(\mathbf{x}; \boldsymbol{\theta}_{t-1}) }{\nu_t \sigma_t(\mathbf{x})} >   \frac{ \lvert f(\mathbf{x}) - h(\mathbf{x}; \boldsymbol{\theta}_{t-1}) \rvert - \epsilon(m) }{\nu_t \sigma_t(\mathbf{x})}  \Bigg \lvert \mathcal{F}_{t-1} \Bigg) \geq \frac{1}{4e\pi}
    \end{split}
\end{equation*}
\end{pl3}

For any step $t$, we consider how the standard deviation of the estimates for each point is, in comparison with the standard deviation for the optimal value.
Following \cite{zhang2021neural}, we divide the discretization $\mathcal D_t$ into two sets: saturated and unsaturated sets. For more details, we define the set of saturated points as:

\begin{equation}
\label{def:saturated_set}
S_t = \{\forall \mathbf{x} \in \mathcal{D}_t, f([\mathbf{x}^*]_t) - f(\mathbf{x}) \geq (1+c_t)\nu_t \sigma_t(\mathbf{x})+ 2\epsilon(m)\}
\end{equation}




The following lemma shows that, the algorithm can pick unsaturated points $\mathbf{x}_t$ with high probability. 
\begin{lemma}[Lemma 4.5, \cite{zhou2020neural}]
\label{lemma:unsaturated_points}

Let $\mathbf{x}_t$ be the chosen point at step $t \in [T]$. Then, $\mathbb{P}(\mathbf{x}_t \not\in S_t \lvert \mathcal{F}_{t-1}) \geq \frac{1}{4e\sqrt{\pi}} - \frac{1}{t^2}$
\end{lemma}

The next lemma bounds the expectation of instantaneous regret at each round conditioned on history $\mathcal{F}_{t-1}$.

\begin{lemma}
\label{lemma:regret_expectation}
Suppose the width of the neural network m satisfies Condition \ref{cond:set}. Set $\eta = C_1(m\lambda + mLT)^{-1}$. Then with probability at least $1- \alpha$, we have for all $t \in [T]$ that
\begin{equation*}
\mathbb{E}[f(\mathbf{x}^*) - f(\mathbf{x}_t) \lvert \mathcal{F}_{t-1}] \leq C_2 (1+c_t)\nu_t \sqrt{L} \mathbb{E}[ \min(\sigma_t(\mathbf{x}_t),B)\lvert\mathcal{F}_{t-1}] + 4\epsilon(m) + \frac{2B+1}{t^2}
\end{equation*}
where $C_1, C_2$ are absolute constants.
\end{lemma}

\newproof{pl4}{Proof of Lemma \ref{lemma:regret_expectation}}
\begin{pl4}
This proof inherits the proof of Lemma 4.6 \cite{zhang2021neural}, and by using the result of unsaturated point $\mathbf{x}_t$ in Lemma \ref{lemma:unsaturated_points}, along with $\lvert f(x)  \rvert = \lvert \langle f, k(x,\cdot) \rangle \rvert \leq B$ instead of $\lvert f(x) \rvert  \leq 1$ as in \cite{zhang2021neural}, we have the following result:
\begin{equation*}
\begin{split}
    &  \mathbb{E} [f([\mathbf{x}^*]_t) - f(\mathbf{x}_t) \mid \mathcal{F}_{t-1}] \\
    \leq  & 44e\sqrt{\pi}(1+c_t )\nu_t C_1\sqrt {L} \mathbb{E}[\min \{\sigma_t(\mathbf{x}_t),B\} \mid \mathcal{F}_{t-1}] + 4\epsilon(m) + \frac{2B}{t^2}
\end{split}
\end{equation*}
Now using Eqn \ref{eqn:rkhs_lipschitz}, we have the instantaneous regret at round $t$
\[
r_t = f(\mathbf{x}^*) - f([\mathbf{x}^*]_t) + f([\mathbf{x}^*]_t) - f(\mathbf{x}_t) \leq \frac{1}{t^2} + f([\mathbf{x}^*]_t) - f(\mathbf{x}_t)
\]
Taking conditional expectation, we have the result as stated in Lemma \ref{lemma:regret_expectation}.
\end{pl4}

The next lemma bounds the cumulative regret $\sum_{t=1}^T r_t$ after $T$ iterations. 
\begin{lemma}
\label{lemma:regret_bound}
Suppose the width of the neural network m satisfies Condition \ref{cond:set}. Then set $\eta = C_1(m\lambda + mLT))^{-1}$, we have, with probability at least $1-\alpha$, that
\begin{equation*}
\begin{split}
   \sum_{t=1}^T f(\mathbf{x^*}) - f(\mathbf{x}_t)
    & \leq  4T \epsilon(m) + \frac{(2B+1)\pi^2}{6} + C_2(1+c_T)\nu_t \sum^T_{t=1} \min(\sigma_t(\mathbf{x}_t),B) \\ &
    + (4B+ C_3 (1+c_T)\nu_t L + 4\epsilon(m)) \sqrt{2 \log(1/\alpha)T}
\end{split}
\end{equation*}
where $C_1, C_2, C_3$ are absolute constants.
\end{lemma}
\newproof{pl5}{Proof of Lemma \ref{lemma:regret_bound}}
\begin{pl5}
Similar to Lemma \ref{lemma:regret_expectation}, we utilize the proof of Lemma 4.7 in \cite{zhang2021neural} or equivalent proof of Lemma 13 in \cite{chowdhury2017kernelized}. Then with $f(\mathbf{x})\leq B$, we have

\begin{equation*}
\begin{split}
    \sum_{i=1}^T r_t & \leq 4T\epsilon(m) + (2B+1) \sum_{i=1}^T t^{-2} + C_1 (1+c_T)\nu_t\sum_{i=1}^T \min(\sigma_t(\mathbf{x}_t),B) \\ & + (4B + C_1C_2(1 + c_T )\nu_t L + 4\epsilon(m))\sqrt{2 \log(1/\alpha)T}
\end{split} 
\end{equation*}
\end{pl5}

The next lemma gives a bound on the sum of variance $\sum_{i=1}^T \min(\sigma_t(\mathbf{x}_t), B)$ which appears in Lemma \ref{lemma:regret_bound}.
\begin{lemma}
\label{lemma:min_sigma}
Suppose the width of the neural network m satisfies Condition \ref{cond:set}. Then set $\eta = C_1(m\lambda + mLT))^{-1}$, we have, with probability at least $1-\alpha$, that
\begin{equation*}
    \sum_{i=1}^T \min(\sigma_t(\mathbf{x}_t),B) \leq \sqrt{\frac{\lambda BT}{\log(B+1)} (2\gamma_T+1) } 
\end{equation*}
\end{lemma}
To prove lemma \ref{lemma:min_sigma}, we first need to utilize a technical lemma:
\begin{sublemma}
\label{lemma:min_B_vs_cov_norm}
Let $\{ \mathbf{v}_t\}_{t=1}^\infty$ be a sequence in $\mathbb{R}^p$, and
define $\mathbf{V}_t = \lambda \mathbf{I} + \sum^t_{i=1} \mathbf{v}_i\mathbf{v}_i^\top$ and $B$ is a positive constant. If $\lambda \geq 1$, then

\[ 
\sum_{i=1}^T \min \{ \mathbf{v}_t^\top \mathbf{V}_{t-1}^{-1}\mathbf{v}_{t-1}, B\} \leq \frac{B}{\log(B+1)} \log \det(\mathbf{I} + \lambda^{-1}\sum_{i=1}^T \mathbf{v}_i\mathbf{v}_i^\top)
\]
\end{sublemma}
\ref{min_B_vs_cov_norm_proof} provided the proof for Lemma \ref{lemma:min_B_vs_cov_norm}. Now, we start to prove Lemma \ref{lemma:min_sigma}
\newproof{pl6}{Proof of Lemma \ref{lemma:min_sigma}}
\begin{pl6}

From Cauchy-Schwartz inequality, we have 
\begin{equation*}
    \begin{split}
        \sum_{i=1}^T \min \{\sigma_t(\mathbf{x}_t), B\} & \leq \sqrt{T \sum_{i=1}^T \min \{\sigma_t^2(\mathbf{x}_t), B\}} \\
    \end{split}
\end{equation*}
We also have, 
\begin{equation*}
    \begin{split}
        \sum_{i=1}^T \min \{\sigma_t^2 (\mathbf{x}_t), B\} & \leq \lambda \sum_{i=1}^T \min \{ \mathbf{g}(\mathbf{x}_t;\boldsymbol{\theta}_0)^\top \mathbf{U}^{-1}_{t-1} \mathbf{g}(\mathbf{x}_t;\boldsymbol{\theta}_0)/m,B \} 
        \\
        & \leq \frac{\lambda B}{\log(B+1)}  \log \det (\mathbf{I} + \lambda^{-1} \sum_{i=1}^T \mathbf{g}(\mathbf{x}_t;\boldsymbol{\theta}_0) \mathbf{g}(\mathbf{x}_t;\boldsymbol{\theta}_0)^\top /m) \\
        & = \frac{\lambda B}{\log(B+1)}  \log \det (\mathbf{I}  + \lambda^{-1} \mathbf{G}_T \mathbf{G}_T^\top /m ) \\
        & = \frac{\lambda B}{\log(B+1)}\log \det (\mathbf{I}  + \lambda^{-1} \mathbf{G}_T^\top \mathbf{G}_T /m ) \\
        & = \frac{\lambda B}{\log(B+1)} \log \det (\mathbf{I}  + \lambda^{-1} \mathbf{K}_T)\\
        & \leq  \frac{\lambda B}{\log(B+1)} (2\gamma_T+1)
    \end{split}
\end{equation*}
where the first inequality moves the positive parameter $\lambda$ outside the min operator. Then the second inequality utilizes Lemma \ref{lemma:min_B_vs_cov_norm}, the first equality use the expression of $\mathbf{G}_t$ in \ref{alg:NeuralBO}, the second equality is from the fact that $\det(\mathbf{I} + \mathbf{A}\mathbf{A}^\top) = \det(\mathbf{I} + \mathbf{A}^\top\mathbf{A})$, and the last equality uses the definition of $\mathbf{K}_T$ in Definition \ref{def:linear_kernelized_terms} and the last inequality uses Lemma \ref{lemma:log_det_Kt_bound}. Finally, we have, 
\[
\sum_{i=1}^T \min \{\sigma_t(\mathbf{x}_t), B\} \leq \sqrt{\frac{\lambda BT}{\log(B+1)} (2\gamma_T+1)}.
\]
\end{pl6}

\textbf{Finally, we repeat to emphasize the proof of Theorem \ref{theorem:main} given in Section \ref{sec:regret_analysis}.}
\newproof{pthmapp}{Proof of Theorem \ref{theorem:main}}
\begin{pthmapp}
\label{proof:theorem_main_appendix}
With probability at least $1-\alpha$, we have
\begin{align*}
 R_T &  = \sum^T_{t=1} f(\mathbf{x^*}) - f(\mathbf{x}_t) \\ 
     & = \sum^T_{t=1} \left[f(\mathbf{x}^*) - f([\mathbf{x}^*]_t)\right] + \left[f([\mathbf{x}^*]_t) - f(\mathbf{x}_t) \right] \\ 
     &\leq 4T \epsilon(m) + \frac{(2B+1)\pi^2}{6} + \Bar{C_1}(1+c_T)\nu_T \sqrt{L} \sum^T_{i=1} \min(\sigma_t(\mathbf{x}_t), B) \\
     & \quad +(4+\Bar{C_2}(1+c_T)\nu_T L + 4\epsilon(m))\sqrt{2 \log(1/\alpha)T} \\
     & \leq \Bar{C_1}(1+c_T)\nu_T \sqrt{L} \sqrt{\frac{\lambda BT}{\log(B+1)} (2\gamma_T+1)} 
     + \frac{(2B+1)\pi^2}{6} + 4T \epsilon(m) \\
     & \quad + 4\epsilon(m)\sqrt{2 \log(1/\alpha)T}  +  \left(4+\Bar{C_2}(1+c_T)\nu_T L\right)\sqrt{2 \log(1/\alpha)T}  \\
     &  = \Bar{C_1}(1+c_T)\nu_t \sqrt{L} \sqrt{\frac{\lambda BT}{\log(B+1)} (2\gamma_T+1)} + \frac{(2B+1)\pi^2}{6} \\
     & \quad +  \epsilon(m)(4T+ \sqrt{2 \log(1/\alpha)T}) + (4+\Bar{C_2}(1+c_T)\nu_t L)\sqrt{2 \log(1/\alpha)T} \\
     \end{align*} 
The first inequality is due to Lemma \ref{lemma:regret_bound}, which provides the bound for cumulative regret $R_T$ in terms of $\sum^T_{t=1} \min(\sigma_t(\mathbf{x}_t),B)$.  The second inequality further provides the bound of term $\sum^T_{t=1} \min(\sigma_t(\mathbf{x}_t),B)$ due to Lemma \ref{lemma:min_sigma}, while the last equality rearranges addition.   Picking $\eta = (m\lambda + mLT)^{-1}$ and $J = \left(1+LT/\lambda \right) \left(\log (C_{\epsilon,2} ) + \log(T^3L\lambda^{-1}\log(1/\alpha)) \right)$, we have 
\begin{equation*}
\begin{split}
      &\frac{b}{a} C_{\epsilon,2}(1 - \eta m \lambda)^J \sqrt{TL/\lambda} \left(4T+\sqrt{2 \log(1/\alpha)T}\right)\\
    = & \frac{b}{a} C_{\epsilon,2} \left(1-\frac{1}{1+LT/\lambda}\right)^{J} \left(4T+\sqrt{2 \log(1/\alpha)T}\right) \\
    = & \frac{b}{a} C_{\epsilon,2} e^{-\left(\log \left(C_{\epsilon,2}\right) + \log(T^3L\lambda^{-1}\log(1/\alpha)) \right)} \left(4T+\sqrt{2 \log(1/\alpha)T}\right)\\
    = & \frac{b}{a}  \frac{1}{C_{\epsilon,2}}.T^{-3}L^{-1}\lambda \log^{-1}(1/\alpha) \left(4T+\sqrt{2 \log(1/\alpha)T}\right)  \le   \frac{b}{a}\\
\end{split}
\end{equation*}
Then choosing $m$ that satisfies:
\begin{equation*}
    \begin{split}
        \left(\frac{b}{a} C_{\epsilon,1} m^{-1/6}\lambda^{-2/3}L^3 \sqrt{\log m} + \left(\frac{b}{a}\right)^3 C_{\epsilon,3} m^{-1/6} \sqrt{\log m} L^4 T^{5/3} \lambda^{-5/3} (1+\sqrt{T/\lambda})\right) \\
        \left(4T+  \sqrt{2 \log(1/\alpha)T}\right) \le \left(\frac{b}{a}\right)^3 
    \end{split}
\end{equation*}
We finally achieve the bound of $R_T$ as:
\begin{flalign*}
R_T & \leq \Bar C_1(1+c_T)\nu_T \sqrt{L} \sqrt{\frac{\lambda BT}{\log(B+1)} (2\gamma_T+1)}  \\
     & +  (4+ \bar C_2(1+c_T)\nu_T L)\sqrt{2 \log(1/\alpha)T} + \frac{(2B+1)\pi^2}{6} + \frac{b(a^2+b^2)}{a^3}
\end{flalign*}
\end{pthmapp}
\section{Proof of Auxiliary Lemmas}
\label{sec:aux_appendix}
\subsection{Proof of Lemma \ref{lemma:NN_vs_linear}}
\label{NN_vs_linear_proof}


We strictly inherit the technique used in Lemma 4.1 \cite{cao2019generalization}, Lemma B.3 \cite{cao2019generalization} and Theorem 5 \cite{allen2019convergence}, combine with Lemma \ref{lemma:bound_hil} to achieve following results:
\begin{subsublemma}
\label{lemma:NTK_related_bounds}
There exists positive constants $\{C_i\}^2_{i=1}$ and $\Bar{C}_{\epsilon,1}$ such that for any $\alpha  \in (0,1]$, if $\tau$ satisfies that:
\[C_{1}m^{-3/2}L^{-3/2}\left[\log(T L^2/\alpha)\right]^{3/2}\leq\tau\leq C_{2}L^{-6}[\log m]^{-3/2},\] then with probability at least $1-\alpha$ over the randomness of $\boldsymbol{\theta}_0$ and $\mathbf{x} \in \{\mathbf{x}_1, \dots \mathbf{x}_T\}$ with $0<  a \le \norm{\mathbf{x}}_2 \le b$:
\begin{enumerate}
    \item For all  $\boldsymbol{\Hat{\theta}}, \boldsymbol{\widetilde{\theta}}$  satisfying $\norm{\boldsymbol{\theta}_0 -  \boldsymbol{\Hat{\theta}}}_2 \le \tau$, $\norm{\boldsymbol{\theta}_0 -  \boldsymbol{\widetilde{\theta}}}_2 \leq \tau$, 
    \[\left \lvert h(\mathbf{x};\boldsymbol{\Hat \theta}) - h(\mathbf{x}; \boldsymbol{\widetilde{\theta}}) - \langle \mathbf{g}(\mathbf{x}; \boldsymbol{\theta}_0), \boldsymbol{\Hat \theta} - \boldsymbol{\widetilde{\theta}} \rangle  \right\rvert  \leq  \frac{b}{a} \Bar{C}_{\epsilon,1}   \tau^{4/3}L^{3}\sqrt{m\log m}\]
    \label{res:true_f_vs_linear}
    
    \item For all  $\boldsymbol{\theta}$  satisfying $\norm{\boldsymbol{\theta}_0 -  \boldsymbol{\theta}}_2 \le \tau$, 
    \[ \norm{\mathbf{g}(\mathbf{x}; \boldsymbol{\theta}) - \mathbf{g}(\mathbf{x}; \boldsymbol{\theta}_0)}_2 \leq \frac{b}{a} \Bar{C}_{\epsilon,1} \sqrt{\log m}\tau^{1/3}L^{3} \norm{\mathbf{g}(\mathbf{x};\boldsymbol{\theta}_0)}_2\]
    
    \label{res:grad_diff_bound}
    
    \item For all  $\boldsymbol{\theta}$  satisfying $\norm{\boldsymbol{\theta}_0 -  \boldsymbol{\theta}}_2 \le \tau$, 
    \[\norm{\mathbf{g}(\mathbf{x};\boldsymbol{\theta})}_F \leq \frac{b}{a} \Bar{C}_{\epsilon,1} \sqrt{mL}\] 
    
    \label{res:grad_bound}
\end{enumerate}
\end{subsublemma}
The next lemma controls the difference between the
parameter of neural network learned by gradient descent and the theoretical optimal solution of the linearized network.

\begin{subsublemma}
\label{lemma:lin_vs_regresion}
There exists constants $\{C_i\}^4_{i=1}$ and $\Bar{C}_{\epsilon,2}$ such that for
any $\alpha \in (0,1]$, if $\eta, m$ satisfy that for all $t \in [T]$ 

\begin{equation*}
\begin{split}
    & 2\sqrt{t/\lambda}\ge C_{1}m^{-1}L^{-3/2}\left[\log(T L^{2}/\alpha)\right]^{3/2}, \\
    & 2\sqrt{t/\lambda}\leq C_{2}\operatorname*{min}\left\{m^{1/2}L^{-6}[\log m]^{-3/2},m^{7/8}(\lambda^2 \eta^2L^{-6}t^{-1}(\log m)^{-1}\right\},\\
    & \eta\le C_{3}(m\lambda+t m L)^{-1}, \\
    & m^{1/6}\ge C_{4}\sqrt{\log m}L^{7/2}t^{7/6}\lambda^{-7/6}(1\ +\sqrt{t/\lambda}), \\
\end{split}
\end{equation*}
then with probability at least $1-\alpha$ over the randomness of $\boldsymbol{\theta}_0$, $\mathbf{x} \in \{\mathbf{x}_1, \dots \mathbf{x}_T\}$ with $0<  a \le \norm{\mathbf{x}}_2 \le b$, we have $\norm{\boldsymbol{\theta}_0 -  \boldsymbol{\theta}}_2 \le 2{\sqrt{t/(m\lambda)}}$ and
    \begin{equation*}
        \begin{split}
            &\norm{\boldsymbol{\theta}_{t-1} - \boldsymbol{\theta}_0 - \mathbf{U}^{-1}_{t-1}\mathbf{G}_{t-1} \mathbf{r}_{t-1}/m}_2\\
         & \leq(1-\eta m\lambda)^{J}\sqrt{t/(m\lambda)}+\left(\frac{b}{a}\right)^2 \Bar{C}_{\epsilon, 2}m^{-2/3}\sqrt{\log m}L^{7/2}t^{5/3}\lambda^{-5/3}(1+\sqrt{t/\lambda}) \\
        \end{split}
    \end{equation*}
\end{subsublemma}

\newproof{pal1}{Proof of Lemma \ref{lemma:NN_vs_linear}}
\begin{pal1}
Set $\tau=2\sqrt{t/m\lambda}$, it can be verified that $\tau$ satisfies condition in \ref{lemma:NTK_related_bounds}. Therefore,  by inequality \ref{res:true_f_vs_linear} in Lemma \ref{lemma:NTK_related_bounds}, and with the initialization of the network $f(\mathbf{x};\boldsymbol{\theta}_0) = 0$, there exists a constant $C_{\epsilon,1}$ such that
\[\left \lvert h(\mathbf{x}; \boldsymbol{\theta}_{t-1}) - \langle \mathbf{g}(\mathbf{x}; \boldsymbol{\theta}_0), \boldsymbol{\theta}_{t-1} - \boldsymbol{\theta}_0 \rangle  \right\rvert  \leq  \frac{b}{a} C_{\epsilon,1}   t^{2/3}m^{-1/6}\lambda^{-2/3} L^{3}\sqrt{m\log m} \]
Then, we have the difference between linearized model and theoretical optimal solution of kernelized regression
\begin{equation*}
    \begin{split}
        & \lvert \langle\mathbf{g}(\mathbf{x}_{t};\boldsymbol{\theta}_{0}),\boldsymbol{\theta}_{t-1}-\boldsymbol{\theta}_{0}\rangle-\langle\mathbf{g}(\mathbf{x}_{t};\boldsymbol{\theta}_{0}),\mathbf{U}_{t-1}^{-1}\mathbf{G}_{t-1}\mathbf{r}_{t-1}/m\rangle \rvert \\
        & \leq \norm{\mathbf{g}(\mathbf{x}_t;\boldsymbol{\theta}_0)}_2 \norm{\boldsymbol{\theta}_{t-1}-\boldsymbol{\theta}_{0} - \mathbf{U}_{t-1}^{-1}\mathbf{G}_{t-1}\mathbf{r}_{t-1}/m}_2\\
        & \le \frac{b}{a} \Bar{C}_{\epsilon,1} \sqrt{mL} \bigg((1-\eta m\lambda)^{J}\sqrt{t/(m\lambda)} \\ & \;\;\;\; +\left(\frac{b}{a}\right)^2 \Bar{C}_{\epsilon, 2}m^{-2/3}\sqrt{\log m}L^{7/2}t^{5/3}\lambda^{-5/3}(1+\sqrt{t/\lambda}) \bigg)\\
        & \le \frac{b}{a} C_{\epsilon,2} (1-\eta m\lambda)^{J} \sqrt{tL/(\lambda)} +  \left(\frac{b}{a} \right)^3 C_{\epsilon,3} m^{-1/6} \sqrt{\log m} L^4 t^{5/3} \lambda ^{-5/3} \left(1 + \sqrt{t/\lambda} \right),\\
    \end{split}
\end{equation*}
where the first inequality is from (\ref{res:grad_bound}) and Lemma \ref{lemma:lin_vs_regresion}, with $C_{\epsilon,2} = \bar C_{\epsilon,1}$ and  $C_{\epsilon,3} = \bar C_{\epsilon,1} \bar C_{\epsilon,2}$.
Finally, we have 
\begin{equation*}
    \begin{split}
        & \left \lvert h(\mathbf{x}; \boldsymbol{\theta}_{t-1}) - \langle\mathbf{g}(\mathbf{x}_{t};\boldsymbol{\theta}_{0}),\mathbf{U}_{t-1}^{-1}\mathbf{G}_{t-1}\mathbf{r}_{t-1}/m\rangle \right\rvert \\
        & \leq  \left \lvert h(\mathbf{x}; \boldsymbol{\theta}_{t-1}) - \langle \mathbf{g}(\mathbf{x}; \boldsymbol{\theta}_0), \boldsymbol{\theta}_{t-1} - \boldsymbol{\theta}_0 \rangle  \right\rvert \\ 
        &  + \lvert \langle\mathbf{g}(\mathbf{x}_{t};\boldsymbol{\theta}_{0}),\boldsymbol{\theta}_{t-1}-\boldsymbol{\theta}_{0}\rangle-\langle\mathbf{g}(\mathbf{x}_{t};\boldsymbol{\theta}_{0}),\mathbf{U}_{t-1}^{-1}\mathbf{G}_{t-1}\mathbf{r}_{t-1}/m\rangle \rvert \\
        & \leq \frac{b}{a} C_{\epsilon,1}   t^{2/3}m^{-1/6}\lambda^{-2/3} L^{3}\sqrt{m\log m} + \frac{b}{a} C_{\epsilon,2} (1-\eta m\lambda)^{J} \sqrt{tL/(\lambda)} \\  
        & +\left(\frac{b}{a} \right)^3 C_{\epsilon,3} m^{-1/6} \sqrt{\log m} L^4 t^{5/3} \lambda ^{-5/3} \left(1 + \sqrt{t/\lambda} \right)  
    \end{split}
\end{equation*} 
as stated in Lemma \ref{lemma:NN_vs_linear}.
\end{pal1}

\subsection{Proof of Lemma \ref{lemma:noise_affeted_bound}}
\label{noise_affeted_bound_proof}
\newproof{pal4}{Proof of Lemma \ref{lemma:noise_affeted_bound}}

\begin{pal4}

We bound the noise-affected term using the sub-Gaussianity assumption. Let $\mathbf{Z}_{t-1}^\top(\mathbf{x})  = \mathbf{g}(\mathbf{x}; \boldsymbol{\theta}_0)^\top \mathbf{U}^{-1}_{t-1} \mathbf{G}_{t-1} /m$. We will show that $\mathbf{Z}_{t-1}^\top(\mathbf{x}) \bold{\epsilon}_{t-1}$ is a sub-Gaussian random variable whose moment generating function is bounded by that
of a Gaussian random variable with variance $\frac{R^2\sigma_{t}^2(\mathbf{x})}{\lambda}$.

Following the proof style in \cite{vakili2021optimal}, we bound the noise-affected term $\mathbf{g}(\mathbf{x}; \boldsymbol{\theta}_0)^\top \mathbf{U}^{-1}_{t-1} \mathbf{G}_{t-1} \boldsymbol{\epsilon}_{t-1}/m = \mathbf{Z}_{t-1}^\top (\mathbf{x})\boldsymbol{\epsilon}_{t-1}$. Let $\zeta_i(\mathbf{x}) = [\mathbf{Z}_{t-1}(\mathbf{x})]_i$, then we have: 
\begin{equation}
\begin{split}
    \mathbb E \left[\exp ( \mathbf{Z}_{t-1}^\top(\mathbf{x})\boldsymbol{\epsilon}_{t-1} ) \right] & =  \mathbb{E}\left[ \exp \sum_{i=1}^{t-1} \zeta_i(\mathbf{x}) \boldsymbol{\epsilon}_i  \right] \\
    & = \prod_{i=1}^{t-1} \mathbb{E} [\exp \zeta_i(\mathbf{x}) \boldsymbol{\epsilon}_i] \\
    &  \leq \prod_{i=1}^{t-1} \exp \left( \frac{R^2 \zeta_i^2(\mathbf{x})}{2} \right)\\
    &  = \exp \left (\frac{R^2 \sum_{i=1}^{t-1}\zeta_i^2(\mathbf{x})}{2} \right) \\
     & = \mathbb \exp \left( \frac{R^2\norm{\mathbf{Z}_{t-1}(\mathbf{x})}^2_2}{2} \right), \\
\end{split}
\label{eqn:noise_effect}
\end{equation}
where the second equation is the consequence of independence of $\zeta_i(\mathbf{x})\boldsymbol{\epsilon}_i$, which directly utilizes our i.i.d noise assumption which is mentioned in Assumption \ref{assumption:iid_noise}. The first inequality holds by the concentration property of the sub-Gaussian noise random variable with parameter $R$. Then we bound:  
\begin{equation}
\label{eqn:bound_Z}
\begin{split}
        \norm{\mathbf{Z}_{t-1}(\mathbf{x})}^2_2 & = \mathbf{Z}_{t-1}(\mathbf{x}) \mathbf{Z}_{t-1}^\top(\mathbf{x})\\
        & = \frac{1}{m^2} \mathbf{g}(\mathbf{x}; \boldsymbol{\theta}_0)^\top \mathbf{U}^{-1}_{t-1} \mathbf{G}_{t-1} \mathbf{G}_{t-1}^\top \mathbf{U}^{-1}_{t-1}\mathbf{g}(\mathbf{x}; \boldsymbol{\theta}_0)
        \\
        & = \frac{1}{m} \mathbf{g}(\mathbf{x}; \boldsymbol{\theta}_0)^\top \mathbf{U}^{-1}_{t-1} (\mathbf{U}_{t-1} - \lambda\mathbf{I}) \mathbf{U}^{-1}_{t-1}\mathbf{g}(\mathbf{x}; \boldsymbol{\theta}_0) \\
        & = \frac{1}{m}\mathbf{g}(\mathbf{x}; \boldsymbol{\theta}_0)^\top  (\mathbf{I} - \lambda\mathbf{U}^{-1}_{t-1}) \mathbf{U}^{-1}_{t-1} \mathbf{g}(\mathbf{x}; \boldsymbol{\theta}_0)\\
        & = \frac{1}{m} \mathbf{g}(\mathbf{x}; \boldsymbol{\theta}_0)^\top \left[ \mathbf{U}^{-1}_{t-1} \mathbf{g}(\mathbf{x}; \boldsymbol{\theta}_0) - \lambda \mathbf{U}^{-2}_{t-1}\mathbf{g}(\mathbf{x}; \boldsymbol{\theta}_0) \right] \\ 
        & = \frac{1}{m} \mathbf{g}(\mathbf{x}; \boldsymbol{\theta}_0)^\top \mathbf{U}^{-1}_{t-1} \mathbf{g}(\mathbf{x}; \boldsymbol{\theta}_0) - \frac{\lambda}{m} \mathbf{g}(\mathbf{x}; \boldsymbol{\theta}_0)^\top \mathbf{U}^{-2}_{t-1}\mathbf{g}(\mathbf{x}; \boldsymbol{\theta}_0)\\
        & = \frac{1}{\lambda}\sigma_t^2(\mathbf{x}) - \frac{\lambda}{m} \mathbf{g}(\mathbf{x}; \boldsymbol{\theta}_0)^\top \mathbf{U}^{-2}_{t-1}\mathbf{g}(\mathbf{x}; \boldsymbol{\theta}_0)\\ 
        & = \frac{1}{\lambda}\sigma_t^2(\mathbf{x}) - \frac{\lambda}{m} \norm{\mathbf{g}(\mathbf{x}; \boldsymbol{\theta}_0) \mathbf{U}^{-2}_{t-1}}^2_2 \\
        & \leq \frac{1}{\lambda} \sigma_t^2(\mathbf{x}),
\end{split}
\end{equation}
where the second equality is from definition of $\mathbf{Z}_{t-1}$ and the third inequality is from the note of Definition \ref{def:linear_kernelized_terms}.  The inequality is from the fact that  $\frac{\lambda}{m} \norm{\mathbf{g}(\mathbf{x}; \boldsymbol{\theta}_0) \mathbf{U}^{-2}_{t-1}}^2 \ge 0, \forall \mathbf{x},t$. Replace inequality \ref{eqn:bound_Z} to equation \ref{eqn:noise_effect} then we have 
\[ \mathbb E \left[\exp \left( \mathbf{Z}_{t-1}^\top(\mathbf{x})\boldsymbol{\epsilon}_{t-1}   \right) \right] \leq \exp (\frac{R^2\sigma_t^2(\mathbf{x})}{2\lambda}) \]
Thus, using Chernoff-Hoeffding inequality \cite{antonini2008convergence}, we have with probabilty at least $1-\alpha$: 

\begin{equation*}
    \begin{split}
        \left \lvert \mathbf{Z}_{t-1}^\top(\mathbf{x})\boldsymbol{\epsilon}_{t-1} \right \rvert \leq \frac{R\sigma_t(\mathbf{x})}{\sqrt{\lambda}} \sqrt{2 \log(\frac{1}{\alpha})}
    \end{split}
\end{equation*}
\end{pal4}

\subsection{Proof of Lemma \ref{lemma:log_det_Kt_bound}}
\label{log_det_Kt_bound_proof}
\newproof{pal3}{Proof of Lemma \ref{lemma:log_det_Kt_bound}}
\begin{pal3}
From the definition of $\mathbf{K}_t$ and Lemma B.7 \cite{zhou2020neural}, we have that
\begin{equation*}
\begin{split}
    \log \det(\mathbf{I}+\lambda^{-1}\mathbf{K}_{t})
    & = \log\det \left(\mathbf{I}+\sum_{i=1}^{t}\mathbf{g}({\mathbf{x}_t};\boldsymbol{\theta}_0)\mathbf{g}({\mathbf{x}_t};\boldsymbol{\theta}_0)^\top/(m\lambda)\right) \\
    & = \log \det(\mathbf{I}+\lambda^{-1}\mathbf{H}_t + \lambda^{-1}(\mathbf{K}_t - \mathbf{H}_t))\\
    & \leq \log \det (\mathbf{I}+\lambda^{-1}\mathbf{H}_t)  + \langle (\mathbf{I}+\lambda^{-1}\mathbf{H}_t)^{-1}, \lambda^{-1}(\mathbf{K}_t - \mathbf{H}_t) \rangle \\
    & \leq \log \det (\mathbf{I}+\lambda^{-1}\mathbf{H}_t)  + \norm{(\mathbf{I}+\lambda^{-1}\mathbf{H}_t)^{-1}}_F \norm{ \lambda^{-1}(\mathbf{K}_t - \mathbf{H}_t)}_F \\
    & \leq \log \det (\mathbf{I}+\lambda^{-1}\mathbf{H}_t) + t\sqrt{t} \epsilon \\
    & \leq \log \det (\mathbf{I}+\lambda^{-1}\mathbf{H}_t) + 1 \\
    & \leq 2 \gamma_t + 1, 
\end{split}
\end{equation*}
where the first equality is from the definition of $\mathbf{K}_t$ in Definition \ref{def:linear_kernelized_terms}, the first inequality is from the convexity of $\log \det(\cdot)$ function,
and the second inequality is from the fact that $\langle \mathbf{A}, \mathbf{B} \rangle \le \norm{\mathbf{A}}_F \norm{\mathbf{B}}_F$. The third inequality is
from the fact that $\norm{\mathbf{A}}_F \le \sqrt{t} \norm{\mathbf{A}}_2$ if $\mathbf{A} \in \mathbb{R}^{t \times t}$. The fourth inequality utilizes the choice of $\epsilon = T^{-3/2}$ and the last inequality inherits Lemma 3 in \cite{chowdhury2017kernelized}.
\end{pal3}

\subsection{Proof of Lemma \ref{lemma:min_B_vs_cov_norm}}
\label{min_B_vs_cov_norm_proof}
\newproof{pal2}{Proof of Lemma \ref{lemma:min_B_vs_cov_norm}}
\begin{pal2}
By basic linear algebra, we have 
\begin{equation*}
\begin{split}
    \det (\mathbf{V}_t)  & = \det (\mathbf{V}_{t-1} + \mathbf{v}_n\mathbf{v}_n^\top) = \det(\mathbf{V}_t)\det(\mathbf{I}+\mathbf{V}_t^{-1/2}\mathbf{v}_n(\mathbf{V}_t^{-1/2}\mathbf{v}_n)^\top) \\
  & = \det(\mathbf{V}_{t-1})(1+\norm{\mathbf{v}_{t-1}}^2_{\mathbf{V}_{t-1}^{-1}}) \\
  & = \det (\lambda \mathbf{I}) \prod_{t=1}^T (1+\norm{\mathbf{v}_{t-1}}^2_{\mathbf{V}_{t-1}^{-1}}) \\
\end{split}
\end{equation*}
where we used that all the eigenvalues of a matrix of the form $\mathbf{I} + \mathbf{x}\mathbf{x}^\top$ are 1, except one eigenvalue, which is $1+\norm{\mathbf{x}}^2$ and which corresponds to the eigenvector $\mathbf{x}$. Using $\min\{u,B\} \leq \frac{B}{\log(B+1)}\log(1+u), \forall u \in [0,B]$, we get 
\begin{equation*}
    \begin{split}
        \sum_{t=1}^T \min \{B,\norm{\mathbf{v}_{t-1}}^2_{\mathbf{V}_{t-1}^{-1}}\} & \leq \frac{B}{\log(B+1)} \sum_{t=1}^T \log (1+\norm{\mathbf{v}_{t-1}}^2_{\mathbf{V}_{t-1}^{-1}}) \\
        & \leq \frac{B}{\log(B+1)} \log \det (\mathbf{I} + \lambda^{-1}\sum_{i=1}^T \mathbf{v}_i\mathbf{v}_i^\top )
    \end{split}
\end{equation*}
\end{pal2}







\end{document}